\documentclass[10pt,journal]{IEEEtran}
\usepackage{amsmath,amsfonts}

\usepackage{array}
\usepackage[caption=false,font=normalsize,labelfont=sf,textfont=sf]{subfig}
\usepackage{textcomp}
\usepackage{stfloats}
\usepackage{url}
\usepackage{verbatim}
\usepackage{graphicx}
\usepackage{hyperref}
\usepackage{amsfonts}
\usepackage{multirow}
\usepackage{float}
\usepackage{caption}
\usepackage{booktabs}  
\usepackage{tabularx}  
\usepackage[ruled,vlined]{algorithm2e}
\usepackage{xcolor} 
\ifCLASSOPTIONcompsoc
  \usepackage[nocompress]{cite}
\else
  \usepackage{cite}
\fi
\ifCLASSINFOpdf
\else
\fi
\begin{document}

\title{Content Accuracy and Quality Aware Resource Allocation Based on LP-Guided DRL for ISAC-Driven AIGC Networks}

\author{Ningzhe~Shi,~\IEEEmembership{Graduate Student Member,~IEEE}, Yiqing~Zhou,~\IEEEmembership{Senior Member,~IEEE}, Ling~Liu,~\IEEEmembership{Member,~IEEE}, Jinglin~Shi, Yihao~Wu, Haiwei~Shi, and~Hanxiao~Yu,~\IEEEmembership{Member,~IEEE}
 
\thanks{The work was supported by the National Key Research and Development Program of China granted by No. 2021YFA1000500 and 2021YFA1000501.({\textit{Corresponding authors: Yiqing Zhou.}})

N. Shi, Y. Zhou, L. Liu, J. Shi, Y. Wu, H. Shi, and H. Yu are with the State Key Lab of Processors, Institute of Computing Technology, Chinese Academy of Sciences, Beijing 100190, China, also with the Beijing Key Laboratory of Mobile Computing and Pervasive Device, Beijing 100190, China, and also with the University of Chinese Academy of Sciences, Beijing 100049, China. (e-mail: shiningzhe21b@ict.ac.cn; zhouyiqing@ict.ac.cn; liuling@ict.ac.cn; sjl@ict.ac.cn; wuyihao22z@ict.ac.cn; shihaiwei25@mails.ucas.ac.cn; yuhanxiao@ict.ac.cn)}

}

\markboth{Journal of \LaTeX\ Class Files, January~2026}%
{Shell \MakeLowercase{\textit{et al.}}: Bare Demo of IEEEtran.cls for Computer Society Journals}

\IEEEpubid{\begin{minipage}{\textwidth}\ \centering 1536--1233 \copyright~2026 IEEE. Personal use is permitted, but republication/redistribution requires IEEE permission. See https://www.ieee.org/publications/rights/index.html for more information.
\end{minipage}}

\maketitle

\begin{abstract}
Integrated sensing and communication (ISAC) can enhance artificial intelligence-generated content (AIGC) networks by providing efficient sensing and transmission. Existing AIGC services usually assume that the accuracy of the generated content can be ensured, given accurate input data (e.g., pose image) and command (i.e., prompt), thus only the content generation quality (CGQ) is concerned. However, it is not applicable in ISAC-based AIGC networks, where content generation is based on inaccurate sensed data. Moreover, the AIGC model itself introduces generation errors, which depend on the number of generating steps (i.e., computing resources). Thus, to assess the quality of experience (QoE) of ISAC-based AIGC services, this paper proposes a content accuracy and quality aware service assessment metric (CAQA). Since allocating more resources to sensing and generating improves content accuracy but may reduce communication quality, and vice versa, this sensing-generating (computing)-communication three-dimensional resource tradeoff must be optimized to maximize the average CAQA (AvgCAQA) across all users with AIGC (CAQA-AIGC). This problem is NP-hard, with a large solution space that grows exponentially with the number of users. To solve the CAQA-AIGC problem with low complexity, a standard linear programming (LP) guided deep reinforcement learning (DRL) algorithm with an action filter (LPDRL-F) is proposed. Through the LP-guided approach and the action filter, LPDRL-F can transform the original three-dimensional solution space to two dimensions, reducing complexity while improving the learning performance of DRL. Simulations show that compared to existing DRL and generative diffusion model (GDM) algorithms without LP, LPDRL-F converges faster and finds better resource allocation solutions, thus improving AvgCAQA by more than 10\%. With LPDRL-F, CAQA-AIGC can achieve an improvement in AvgCAQA of more than 50\% compared to existing schemes focusing solely on CGQ.
\end{abstract}


\section{Introduction}\label{sec:introduction}

%
%
%
%
\IEEEpubidadjcol
\IEEEPARstart{W}{ith} the rapid development of data analysis, hardware, and generative artificial intelligence (GAI), AI-generated content (AIGC) is rapidly advancing \cite{ref1,ref2}. It has become one of the key technologies for creating and modifying digital content such as text, images, and videos \cite{ref3,ref4,ref5}. The representative applications of AIGC, such as DALL·E, have shown the ability to generate high-quality and diverse images from scratch, conditioned on a given input \cite{ref4}. These applications rely on traditional human-computer interaction methods, generating content only based on the input text and voice prompts \cite{ref6}. However, some user intents (e.g., physical poses) and environment changes are difficult to express through text and language. Traditional methods capture this information via cameras or sensors, which can be costly and raise privacy concerns \cite{ref7}. For example, extended reality (XR) applications may collect facial features, leading to potential misuse of personal data \cite{ref8}. The emerging integrated sensing and communication (ISAC) can provide a relatively privacy-safe and non-contact alternative by using wireless signals to perceive human poses and environmental information \cite{ref9,ref10,ref11,ref12}. It is promising to integrate AIGC with ISAC to provide various services such as digital human live broadcasts, virtual clothing try-ons, and gaming and film production \cite{ref4,ref13,ref14}.

For instance, a wireless perception-guided AI digital content generation framework is proposed in \cite{ref13}, which optimizes the AIGC service pricing strategy and computing resource allocation to maximize the profit of the AIGC service provider and the utility of users. Similarly, a wireless perception-based AIGC network architecture is proposed in \cite{ref14}, which optimizes computing resource allocation between perception information processing and content generation to enhance AIGC service quality. Nevertheless, existing works \cite{ref13,ref14} mainly focus on computing resource management and incentive mechanism design. A critical gap remains in understanding how sensing and wireless communication resource allocation impact AIGC service performance. Specifically, the joint optimization of sensing, computing, and communication resources and its effects on the quality of AIGC service have not been addressed. Hence, this paper investigates AIGC service provisioning with ISAC resource constraints by proposing a joint sensing, computing, and communication resources optimization framework to enhance the quality of service in ISAC-enabled AIGC networks.
\begin{figure*}[!t]
\centering    
\includegraphics[width=7.0in]{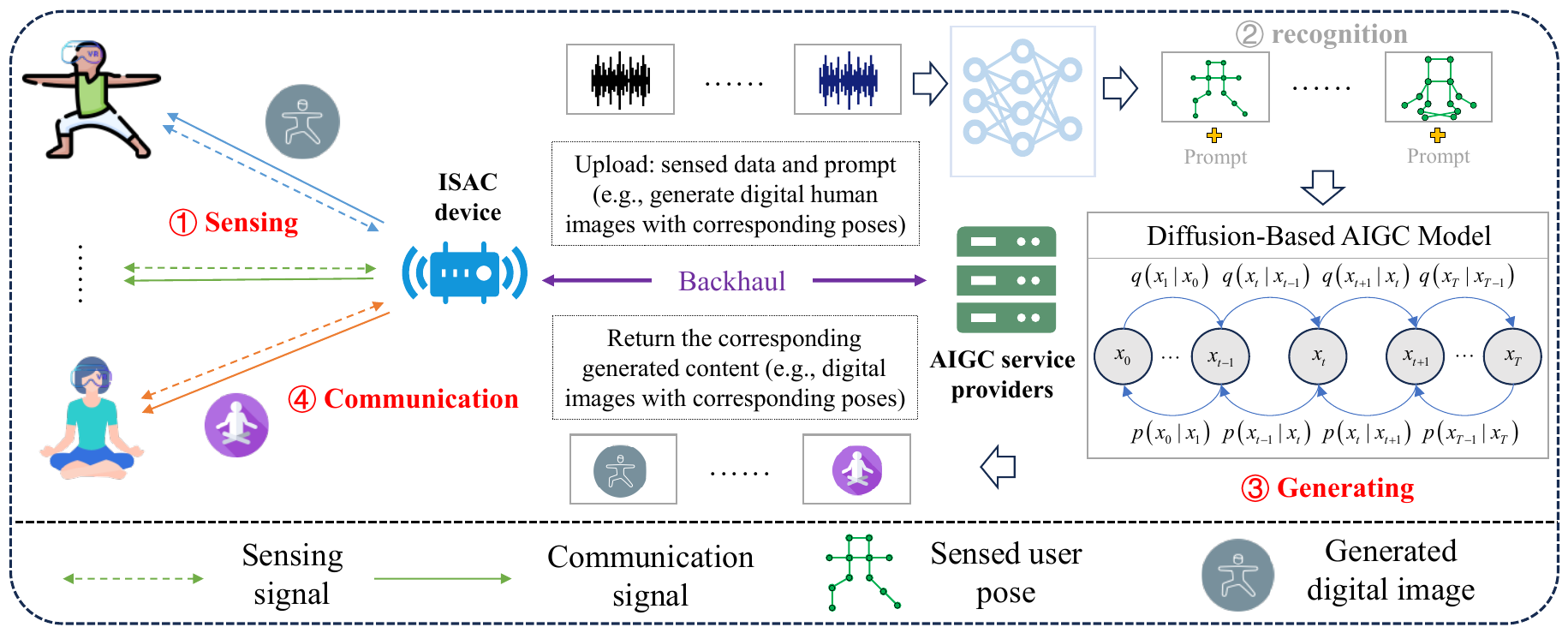}
\caption{An ISAC-based AIGC network.}
\vspace{-1.5em}
\label{Fig1}
\end{figure*}

As shown in Fig. \ref{Fig1}, in an ISAC-based AIGC network, the ISAC device connects to an AIGC service provider via a backhaul link. It can capture the user's pose data via wireless sensing and forward the generated content (e.g., a digital human image with corresponding pose used for virtual avatars in a virtual reality (VR) game) from the AIGC server to the user’s VR device via wireless communication \cite{ref15}. Compared to traditional human-computer interactions, since the sensed pose is used as an input to the AIGC, the sensing accuracy of the ISAC device will affect the precision of generated content at the AIGC server, which in turn influences the quality of the user experience (QoE) during virtual gaming \cite{ref10,ref11}. Meanwhile, the user’s QoE is also influenced by the resolution (quality) of the digital human images generated by AIGC, which depends on the communication capacity. Thus, sensing and communication jointly affect the overall QoE of ISAC-based AIGC service, and both need more resources to enhance the QoE.

Hence, a key challenge in ISAC-based AIGC networks is optimizing ISAC’s resource allocation to maximize QoE. Given that QoE depends on subjective user perceptions, which vary according to individual needs \cite{ref6}, resources should be allocated according to task-oriented differentiated service requirements. Furthermore, existing AIGC services usually assume that the accuracy of the generated content can be ensured, as they generate content based on accurate input data (e.g., pose image) and prompt, thus only the content generation quality (CGQ) is concerned \cite{ref6}. However, it is not applicable in ISAC-based AIGC networks, where input prompts remain accurate, but content generation is based on inaccurate sensed data (the more sensing resources, the more accurate the sensing data). Moreover, the inherent generating accuracy of the AIGC model itself should also be considered, because generating errors will be introduced during the iterative denoising process of AIGC \cite{ref16}. The average error of content generation (AEG) is related to the number of generation steps, model size, and architecture \cite{ref17}. More generation steps result in lower AEG with longer computing time \cite{ref18}.

Therefore, a content accuracy and quality aware service assessment metric is needed, which can evaluate the impact of sensing, generating (computing), and communication resources on the system performance. Based on this metric, an optimization problem can be formulated by jointly optimizing sensing, generating, and communication resource allocation, subject to various constraints. It is shown that this problem is NP-hard, with a large solution space that grows exponentially with the number of users. A smart low complexity solution is needed to solve this problem. The main contributions of this paper are as follows:
\begin{itemize}
\item We propose a novel content accuracy and quality aware service assessment metric (CAQA) for ISAC-based AI-generated content. To maximize the average CAQA (AvgCAQA) across all users with AIGC, an optimization problem is formulated by jointly optimizing sensing, generating, and communication resource allocation (CAQA-AIGC), subject to constraints on the maximum total energy and service time, minimum CAQA requirements, device display capacity, and so on.
\item We propose a low complexity LPDRL-F algorithm to solve this NP-hard problem by decomposing it into two sub-problems, i.e., sensing and generating resource allocation (SGenRA) and communication resource allocation (ComRA). SGenRA is solved by a deep reinforcement learning (DRL) algorithm with an action filter (DRL-F), which can eliminate ineligible sensing resource allocation solutions to reduce the solution space. Based on the results of SGenRA, ComRA becomes a standard linear programming (LP) problem and can be optimally solved.
\item We conduct extensive simulations to verify the effectiveness of the proposed schemes. It will be shown that, compared to existing DRL and generative diffusion model (GDM) algorithms, LPDRL-F converges faster and finds better resource allocation solutions, thus improving AvgCAQA by more than \textbf{10\%}. Moreover, compared to the CGQ-only oriented schemes, the proposed CAQA-AIGC with LPDRL-F improves AvgCAQA by over \textbf{50\%}.
\end{itemize}

The remainder of the paper is organized as follows. Section~\ref{sec:system_model} introduces the system model. The CAQA-AIGC problem of maximizing AvgCAQA is formulated in Section~\ref{sec:AvgCAQA_problem}. Section~\ref{sec:LPDRL-F} details the LPDRL-F algorithm. Section~\ref{sec:performance_evaluation} presents simulation results. Finally, conclusions are drawn in Section~\ref{sec:conclusion}.

\section{System Model}\label{sec:system_model}
\subsection{ISAC-Based AIGC Network Architecture}
As shown in Fig. \ref{Fig1}, consider an ISAC-based multi-user AIGC wireless network with a bandwidth of $B$. Assume a single-antenna ISAC transceiver device, which can switch between the sensing mode and communication mode \cite{ref15}. It connects to the AIGC service provider in the edge server via a backhaul link. $K$ users are randomly distributed within the coverage area of the ISAC device, where each user requests an AIGC task characterized by a task priority level (i.e., sensing order) and a minimum CAQA requirement ${\Omega _{\min }}$. The set of users is denoted as ${\rm{{\cal K}}} = \{ 1, \cdots ,K\} $, where the indices from 1 to $K$ represent users sorted in descending order of task priority level. Notably, due to varying task requirements in ISAC, sensing and communication power may differ \cite{ref13,ref19}.

\begin{figure*}[!t]
\centering    
\includegraphics[width=6.8in]{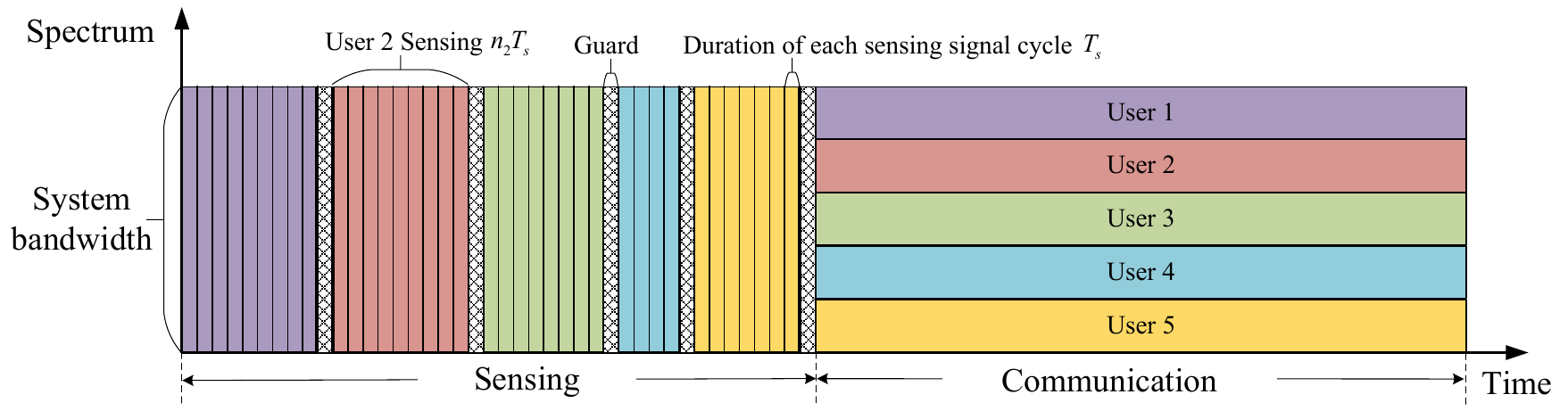}
\caption{Time-frequency diagram of ISAC waveforms.}
\vspace{-1.5em}
\label{Fig2}
\end{figure*}

As shown in Fig. \ref{Fig2}, with the entire bandwidth $B$, the ISAC device first sends frequency-modulated continuous wave (FMCW) to perform pose sensing for $K$ users in a time-division multiplexing (TDM) way according to their priority order, followed by a guarding interval \cite{ref20,ref21}. The sensed pose data are sent to the AIGC provider, which performs pose recognition and generates corresponding digital human images. Assume that the time of pose recognition is negligible compared to that of digital content generation \cite{ref6,ref9,ref15,ref21}. The generated content is then returned to the ISAC device via the backhaul link. Finally, the ISAC device transmits digital human images to the user’s XR devices using orthogonal frequency-division multiplexing (OFDM), where a subchannel of ${B \mathord{\left/
 {\vphantom {B K}} \right.
 \kern-\nulldelimiterspace} K}$ is allocated for each user.

\subsection{ISAC-Based AIGC Service Workflow and Performance Modeling}
\begin{figure*}[!t]
\centering    
\includegraphics[width=7.0in]{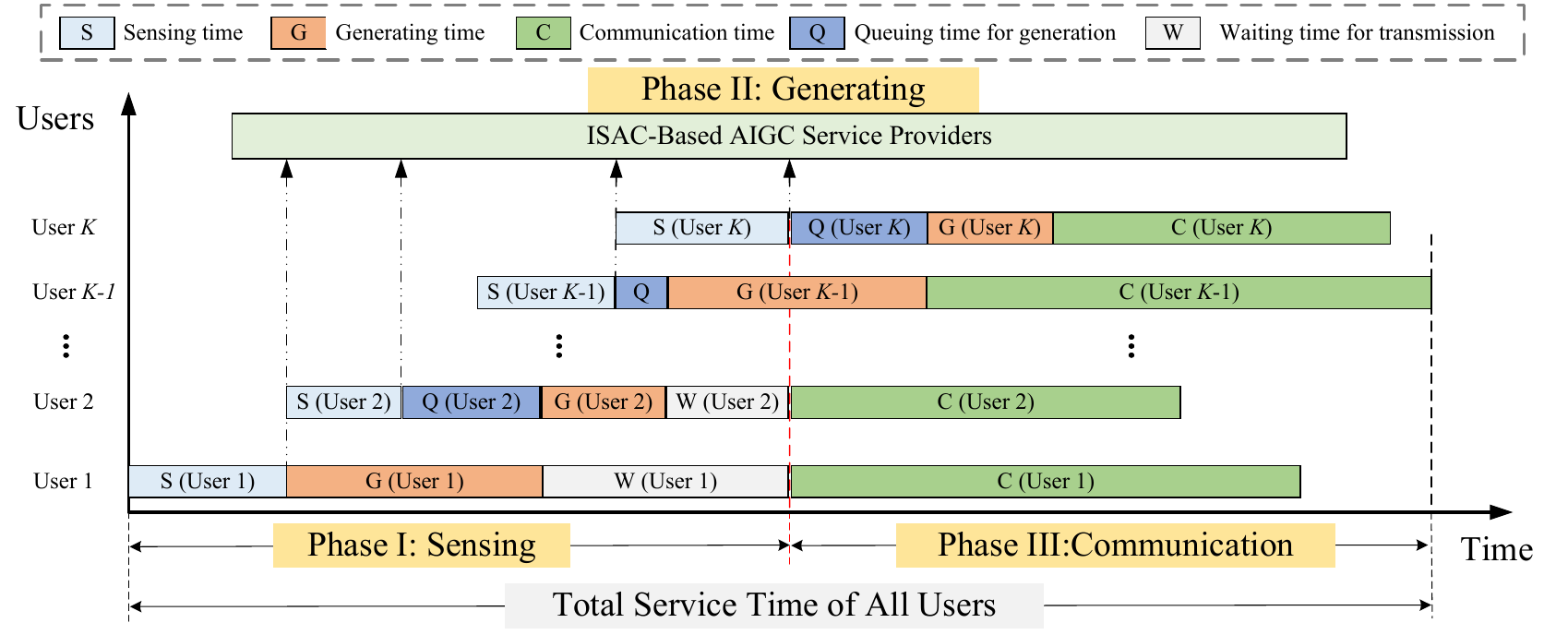}
\caption{Service Process of ISAC-Based AIGC.}
\vspace{-1.5em}
\label{Fig3}
\end{figure*}
Hence, as shown in Fig. \ref{Fig3}, the ISAC-based AIGC network operates in three phases. In Phase I, the ISAC device sends FMCW to perform sensing and obtain the user pose data \cite{ref21}. Let ${n_k}$ denote the number of sensing signal cycles for sensing the $k$-th user. Each cycle of the sensing signal includes a sensing waveform with a duration of ${T_s}$ \cite{ref21,ref22}. The recognition accuracy of user $k$ pose (i.e., sensing accuracy) ${\Upsilon _k}\left( {{n_k}} \right)$ is positively affected by ${n_k}$, given by \cite{ref20,ref21}
\begin{equation}
    {\Upsilon _k}({n_k}) = \xi  - \varpi n_k^{ - \tau },
    \label{Equ1}
\end{equation}
where $\xi \in \left( 0,1 \right]$ is the upper bound of sensing accuracy, determined by hardware and signal characteristics. $\varpi >0$ and $\tau >0$ are parameters to reflect the relationship between sensing cycles and accuracy. $\xi$, $\varpi $, and $\tau$ are usually obtained by fitting empirical sensing accuracy data from practical wireless signal measurements \cite{ref21}. Then the sensing energy of ISAC for user $k$ is given by ${{E}_{k,s}}={{n}_{k}}{{T}_{s}}{{P}_{s}}$, where ${{P}_{s}}$ is the sensing power.

Based on the user pose sensed in Phase I and the prompt, in Phase II, AIGC generates the corresponding digital human image on the server and returns it to the ISAC device via the backhaul link. Assuming sufficient energy resources in the server and large backhaul capacity \cite{ref23}, the time and energy consumption of both upload and return via backhaul are negligible \cite{ref24}. However, the computing capacity of the server is typically limited. Let $f$ denote the computation capacity of the server, measured in Floating Point Operations per second (FLOPS). The generating time of the corresponding digital human image by AIGC is related to the total number of FLOPS, which is proportional to the number of generating steps used by the AIGC \cite{ref17}. Let $\chi $ denote the number of average FLOPS used by one generating step of AIGC (i.e., FLOPS per step). The value of $\chi $ is related to the model size and architecture. Then, the total number of FLOPS used by AIGC to generate user $k$’s content is $\chi  \cdot {z_k}$, where ${z_k}$ is the requested generation steps for user $k$. The generating time of user $k$’s contents is given by ${T_{k,\text{gen}}} = {{\left( {\chi  \cdot {z_k}} \right)} \mathord{\left/
 {\vphantom {{\left( {\chi  \cdot {z_k}} \right)} f}} \right.
 \kern-\nulldelimiterspace} f}$.

Moreover, the average error in the generating phase (AEG) of user $k$ should be concerned, which depends on the number of generation steps ${z_k}$, i.e., the reverse diffusion steps used by the AIGC model for iterative denoising \cite{ref17}. Let ${\varepsilon _k}\left( {{z_k}} \right)$ denote the AEG of user $k$’s content generated by the AIGC model, given by \cite{ref18}
 \begin{equation}
    {\varepsilon _k}\left( {{z_k}} \right) = {\epsilon ^{{\rm{fwd}}}} \cdot {e^{ - \mu  \cdot \left( {{z_k} - {z_{\min }}} \right)}},
    \label{Equ2}
\end{equation}
where ${\epsilon ^{{\rm{fwd}}}} \in \left( {0,1} \right]$ is a scaling factor evaluating the forward process of the AIGC model, $\mu  > 0$ is an attenuation factor characterizing the generating ability of the AIGC model, and ${z_{\min }}$ is the minimum number of generating steps required to obtain a usable generated result.

Assume that the task of generating the corresponding digital human image is processed by the server on a first-come-first-served (FCFS) basis. That is, the server can start processing a newly arrived task only after it has finished processing all previous arrivals. Neglecting propagation time, the task $k$ arrives at the server at time instant ${T_{k,\text{arr}}} = \sum\limits_{i = 1}^k {{n_i}{T_s}}$, and the queuing time of user $k$ is given by
 \begin{equation}
    {T_{k,\text{que}}} = \max \left( {\left( {{T_{k - 1,\text{arr}}} + {T_{k - 1,\text{que}}} + {T_{k - 1,\text{gen}}} - {T_{k,\text{arr}}}} \right),0} \right).
    \label{Equ3}
\end{equation}

In Phase III, the ISAC device transmits the generated images to $K$ users via OFDM. Thus, the waiting time of user $k$ for communication is given by
 \begin{equation}
    {T_{k,\text{wait}}} = \max \left( {\sum\nolimits_{i = 1}^K {{n_i}{T_s}}  - \left( {{T_{k,\text{arr}}} + {T_{k,\text{que}}} + {T_{k,\text{gen}}}} \right),0} \right),
    \label{Equ4}
\end{equation}
where $\sum\nolimits_{i = 1}^K {{n_i}{T_s}}  - \left( {{T_{k,\text{arr}}} + {T_{k,\text{que}}} + {T_{k,\text{gen}}}} \right)$ is the waiting time for communication because even if user $k$ completes its sensing, queuing, and generating processes, it must wait for the ISAC device to finish sensing of all users.

Concerning content quality, note that more communication capacity is needed to transmit a higher quality image. According to Shannon’s theory, the transmission rate of user $k$ is given by
\begin{equation}
    {C_k} = \frac{B}{K}{\log _2}(1 + \frac{{{g_k}{P_c}K}}{{{\delta ^2}B}}),
    \label{Equ5}
\end{equation}
where ${g_k}$ is the channel gain, including large-scale path loss and small-scale Rayleigh fading, ${P_c}$ is the transmission power, and ${\delta ^2}$ is the power spectral density of the additive white Gaussian noise (AWGN). Given that the communication time for user $k$ is ${T_{k,c}}$, the communication capacity is ${C_k}{T_{k,c}}$ and the communication energy of user $k$ is ${E_{k,c}} = {T_{k,c}}{P_c}$. 

Thus, the total completion time of user $k$ includes sensing time, queuing time, generating time, waiting time for communication, and communication time, given by 
\begin{equation}
    {T_k} = {T_{k,\text{arr}}} + {T_{k,\text{que}}} + {T_{k,\text{gen}}} + {T_{k,\text{wait}}} + {T_{k,c}}.
    \label{Equ6}
\end{equation}

In summary, the service quality of ISAC-based AIGC depends on the accuracy and quality of the generated content received by the users. The accuracy of the received content is decided by the sensing accuracy in Phase I and AEG in Phase II, while the quality of the received content is limited by the communication capability in Phase III. As shown in Fig. \ref{Fig3}, sensing and communication share the same radio resources, and AEG is affected by the number of generation steps. Moreover, the sensing, generating, and communication phases are highly coupled through resource consumption (i.e., time and energy) and the queue. Thus, radio and computing (i.e., generation steps) resources should be carefully allocated to sensing, generation, and communication to provide high-quality service.

\section{CAQA-AIGC Problem Formulation}\label{sec:AvgCAQA_problem}
\subsection{QoE Model for ISAC-Based AIGC Service}
Given that QoE depends on subjective user perceptions, which vary according to individual needs \cite{ref6}, resources should be allocated according to task-oriented differentiated service requirements. Furthermore, the service quality of ISAC-based AIGC depends on both the accuracy and quality of the generated content received by the users. A novel QoE model, i.e., CAQA, is proposed, which concerns both the accuracy and quality of generated images. First, since image generation is based on sensed user pose, the generation content accuracy for user $k$ is decided by both the sensing accuracy of ISAC and AEG of the AIGC model, given by
\begin{equation}
    {\Theta _k} = \left( {1 - {\varepsilon _k}} \right) \cdot {\Upsilon _k}\left( {{n_k}} \right),
    \label{Equ7}
\end{equation}
where $\left( {1 - {\varepsilon _k}} \right)$ is the generating accuracy of the AIGC model. 

Second, image quality is positively impacted by resolution \cite{ref25,ref26} until it reaches the maximum display capacity of the user device, and more communication capacity is needed to transmit an image with higher resolution. Assuming that the image is losslessly compressed, and each pixel’s color information occupies a fixed number of bits (e.g., 24-bit RGB representation where each pixel requires 24 bits), the AIGC-generated image resolution, represented by the total number of pixels, is given by ${x_k} = {{\left( {{C_k}{T_{k,c}}} \right)} \mathord{\left/
 {\vphantom {{\left( {{C_k}{T_{k,c}}} \right)} \beta }} \right.
 \kern-\nulldelimiterspace} \beta }$, where $\beta $ is the number of bits per pixel.

Considering that even high image quality is ineffective if the accuracy of the sensing data fed into AIGC is poor, CAQA can be modeled as the product of the accuracy component and quality of the generated image, given by
\begin{equation}
    {\Omega _k} = {\Theta _k}({n_{k}},{z_k}) \cdot \min \left\{ {\frac{{{x_k}}}{{{D_c}}},1} \right\},
    \label{Equ8}
\end{equation}
where ${D_c}$ denotes the user device’s maximum display capacity, and the quality $\min \left\{ {{{{x_k}} \mathord{\left/
{\vphantom {{{x_k}} {{D_c}}}} \right.
\kern-\nulldelimiterspace} {{D_c}}},1} \right\}$ indicates that increasing transmission capacity beyond ${D_c}$ brings no additional quality gain. 

In the CAQA model, content accuracy reflects sensing accuracy and generation errors, which directly impact user-perceived quality because they determine whether the generated result matches the desired one. It primarily influences the visual fidelity of regions of interest (ROI) and desired consistency \cite{zhang2025NeuralVisualQuality,jin2024HeroicVR}, as users are sensitive to whether the desired consistency of ROI regions (e.g., pose consistency) adheres to their intended expectations. Content quality is related to the delivered resolution, which shapes the visual fidelity and fine-grained structural details within ROI regions. Accordingly, content quality mainly corresponds to the psychological factors of visual fidelity of ROI and fine-grained details \cite{zhang2025NeuralVisualQuality,jin2024HeroicVR}. Therefore, a higher CAQA can stand for a better user-perceived quality.

As shown in \eqref{Equ1}, \eqref{Equ5}, \eqref{Equ7}, and ${n_k} = {{{E_{k,s}}} \mathord{\left/
 {\vphantom {{{E_{k,s}}} {\left( {{T_s}{P_s}} \right)}}} \right.
 \kern-\nulldelimiterspace} {\left( {{T_s}{P_s}} \right)}}$, ${\Theta _k}\left( {{n_{k}},{z_k}} \right)$ are positively correlated with sensing energy ${E_{k,s}}$ and generating step $z_k$, and ${x_k}$ is positively correlated with communication energy $E_{k,c}$. Thus, CAQA is jointly influenced by sensing energy, generating step, and communication energy allocation.

\subsection{Problem Formulation}
In the ISAC-based AIGC system, all users share the ISAC device’s energy and spectrum resources and the AIGC server’s computing resources, necessitating a careful allocation of both sensing and communication energy and generating steps to provide good service quality for each user. Therefore, the sensing, generating (computing), and communication resource allocation is optimized for all users with AIGC to maximize AvgCAQA (CAQA-AIGC), given by
\begin{equation}
\begin{aligned}
\text{(CAQA-AIGC):} \quad & \max_{\mathbf{E}_s, \mathbf{z}, \mathbf{E}_c} \ \frac{1}{K} \sum_{k=1}^{K} \Omega_k(E_{k,s}, z_k, E_{k,c}) \\
\text{s.t.} \quad 
& \text{C1: } 0 \le T_k \le T_{\max}, \quad \forall k, \\
& \text{C2: } \sum_{k=1}^{K} E_{k,s} + \sum_{k=1}^{K} E_{k,c} \le E_{\max}, \\
& \text{C3: } 0 \le E_{k,s} \le E_{s,\max}, \quad \forall k, \\
& \text{C4: } z_{\min} \le z_k \le z_{\max}, \quad \forall k, \\
& \text{C5: } \Omega_k \ge \Omega_{k,\min}, \quad \forall k, \\
& \text{C6: } 0 \le x_k \le D_c, \quad \forall k,
\end{aligned}
\label{Equ9}
\end{equation}
where ${{\bf{E}}_s} = \left\{ {{E_{1,s}}, \cdots ,{E_{K,s}}} \right\}$ and ${{\bf{E}}_c} = \left\{ {{E_{1,c}}, \cdots ,{E_{K,c}}} \right\}$ represent the sensing and communication energy for all users, respectively. ${\bf{z}} = \left\{ {{z_1}, \cdots ,{z_K}} \right\}$ represents the generating step for all users. $\text{C1}$ means that the total completion time for any user should be less than the maximum service time ${T_{{\rm{max}}}}$. $\text{C2}$ means that the total energy consumption in the ISAC device should be less than the maximum energy $E_{max}$. $\text{C3}$ means that the sensing energy allocated to each user should not exceed the maximum sensing energy $E_{s,max}$. $\text{C4}$ constrains the generation steps $z_k$ within a feasible range \cite{ref17}. $\text{C5}$ requires that the service quality of each user should meet an individualized minimum CAQA requirement ${\Omega _{k{\rm{,min}}}}$. $\text{C6}$ means that the generated image resolution should be less than the user device display capacity.

The CAQA-AIGC problem is non-convex and NP-hard due to the nonlinear relationship between sensing energy and sensing accuracy, as well as the dependency of AEG on the number of generation steps. Moreover, the resource coupling caused by queuing times for computing during the generation phase (Phase II) and the waiting times for communication before the communication phase (Phase III) jointly affect the service time. This intrinsic coupling significantly increases the complexity of the overall resource allocation problem. Note that given the sensing energy and generating step solutions, the communication energy resource allocation (ComRA) can be transformed into a standard LP problem. However, sensing and generating resource allocation (SGenRA) remains a non-convex and NP-hard problem, which can be well-solved by DRL \cite{ref27}. One possible solution is to decompose the three-dimensional CAQA-AIGC problem into two sub-problems, i.e., the two-dimensional SGenRA sub-problem and the convex ComRA sub-problem. First, SGenRA can be solved by DRL. Then, based on its solution, ComRA can be solved by RCE. Thus, a low complexity LP-guided DRL-F (LPDRL-F) algorithm can be proposed to solve the CAQA-AIGC problem, which reduces the solution space through the LP-guided approach.

\section{LP-Guided DRL-F Algorithm}\label{sec:LPDRL-F}
\subsection{Structure and Main Stages}
\begin{figure*}[!t]
\centering    
\includegraphics[width=6.8in]{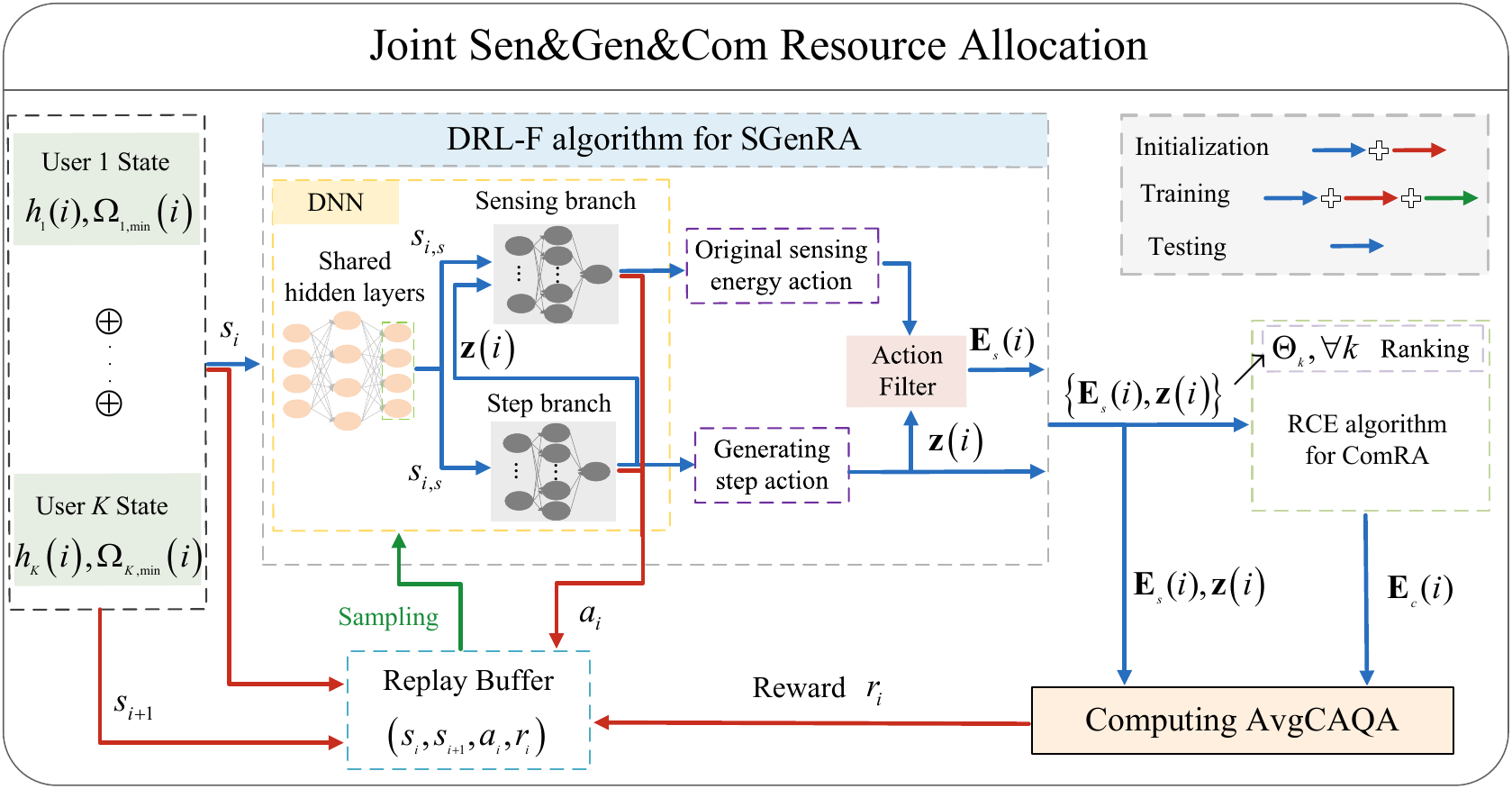}
\caption{Structure of the LP-guided DRL-F Algorithm.}
\vspace{-1.5em}
\label{Fig4}
\end{figure*}
The structure of the proposed LPDRL-F is shown in Fig. \ref{Fig4}. Note that the joint optimization problem is divided into two sub-problems, i.e., SGenRA, and ComRA. First, a DRL-F algorithm is proposed to allocate sensing energy and generating steps based on user states, including channel condition ${h_k}$ and CAQA requirement ${\Omega _{k,\min }}$. Then, with the obtained sensing energy ${{\bf{E}}_s}$ and generating step ${\bf{z}}$, the generating accuracy ${\Theta _k}$ can be obtained. Moreover, ComRA sub-problem can be analytically demonstrated as a standard LP problem, and be optimally solved by a ranking-based communication energy allocation algorithm (RCE). Thus, AvgCAQA can be obtained (see \eqref{Equ8} and \eqref{Equ9}).

Note that in LPDRL-F, the deep neural network (DNN), including shared hidden layers and two separate branches, i.e., the sensing branch and the step branch, should be trained to learn the policy mapping from user states to sensing energy and generating step actions. To enhance the generalization capability of the proposed algorithm, user states with a total size of 2$K$, including $K$ channel conditions and $K$ CAQA requirements, are dynamically changing at each iteration. Specifically, the CAQA requirements follow a uniform distribution \cite{ref17}. At each iteration, user states are input to shared hidden layers of the DNN to extract a shared feature vector ${s_{i,s}}$, which is then fed to two separate branches. Since the generating steps are discrete, Gumbel-softmax method \cite{ref28} is employed to obtain the generating step solutions for $K$ users, i.e., ${\bf{z}}(i)$. These discrete steps are concatenated with the shared features to form the input of the sensing branch. This design enables the sensing branch to capture the feature of the generating steps, thereby ensuring that the output original sensing energy actions are well adapted to the action filter determined by the generating steps. The sensing branch employs a multi-layer perceptron (MLP) with sigmoid activation \cite{ref29} to obtain $K$ continuous values, representing the original sensing energy actions, which are processed by the action filter to obtain sensing energy solutions, i.e., ${{\bf{E}}_s}$. Given joint sensing and generating solutions, i.e., $\left\{ {{{\bf{E}}_s}\left( i \right),{\bf{z}}\left( i \right)} \right\}$, CAQA-AIGC problem can be reduced to ComRA sub-problem, which can be effectively solved by the RCE algorithm with low complexity. Thus, the communication energy allocation solution ${{\bf{E}}_c}$ can be obtained, and the value of AvgCAQA, i.e., $\frac{1}{K}\sum\nolimits_{k = 1}^K {{\Omega _k}\left( {{E_{k,s}},{z_k},{E_{k,c}}} \right)}$ can be calculated and fed back to the DNN. This feedback, user states, and joint sensing energy and generating step action are used as training data to guide the DNN to refine its policy, enabling it to maximize AvgCAQA and optimize the joint sensing, generating, and communication resource allocation strategy over time. Therefore, LPDRL-F involves three stages:

(1)	Initialization Phase (Blue and red arrow flow): The DNN in LPDRL-F is initialized via orthogonal initialization \cite{ref30}, which can effectively mitigate the problem of vanishing or exploding gradients. Note that action filter does not require initialization because it is rule-based (see Section~\ref{subsec:DRL-F}) and determined by the user’s generating steps and CAQA requirements. Similarly, the RCE algorithm also does not require initialization. At the beginning of the algorithm, the replay buffer $\Psi $ is initialized to be empty. Given the observed states ${s_i}$ of all users in the $i$-th iteration, including the channel condition $h_k(i)$ and CAQA requirements ${\Omega _{k,\min }}\left( i \right)$, the joint action ${a_i}$, including the generating step ${\bf{z}}\left( i \right)$ and original sensing energy allocation action, is generated by the DNN. The sensing energy allocation solutions, i.e., ${{\bf{E}}_s}\left( i \right)$, is obtained by the action filter. Then, given ${\bf{z}}\left( i \right)$ and ${{\bf{E}}_s}\left( i \right)$, the communication energy allocation solution ${{\bf{E}}_c}\left( i \right)$ can be obtained by the RCE algorithm. According to \eqref{Equ9}, AvgCAQA is obtained with ${\bf{z}}\left( i \right)$, ${{\bf{E}}_s}\left( i \right)$, and ${{\bf{E}}_c}\left( i \right)$, it is taken as a reward ${r_i}$ for DNN. Then, $\left( {{s_i},{a_i},{r_i},{s_{i + 1}}} \right)$ (see Section~\ref{subsec:DRL-F} for detailed definitions), is input to the replay buffer $\Psi $ as a single experience data. This process continues until the replay buffer accumulates sufficient data to commence training, such as ${\Psi _{\min }} = 5000$ experiences.

(2)	Training Phase (Blue, red, and green arrow flow): Given ${\Psi _{\min }}$ experiences in the replay buffer, DNN in LPDRL-F can be trained. Note that to stabilize the learning process, experience data are randomly drawn with a batch size of ${\Psi_{b}}$, such as ${\Psi_{b}}=256$, from the replay buffer for DNN training \cite{ref29}. As training goes on, the parameters of the DNN in LPDRL-F are progressively optimized, resulting in increasingly effective actions and higher rewards based on observed states. Newly generated experiences continue to be added to the replay buffer, replacing the oldest data. As iterations go on, when the reward demonstrates stability or plateaus, LPDRL-F is considered to have reached convergence. Actions produced after convergence are taken as near-optimal.

(3)	Testing Phase (Blue arrow flow only): After the LPDRL-F algorithm converges, given observed states, DRL-F can directly map them to sensing energy and generation step solutions. Then, given ${{\bf{E}}_s}\left( i \right)$ and ${\bf{z}}\left( i \right)$, the communication energy allocation solution ${{\bf{E}}_c}\left( i \right)$ can be obtained by the RCE algorithm.

Generalization and Retraining: The generalization capability of the trained model is crucial. Although the model is trained on data collected from general scenarios, its performance may degrade when applied to new, unseen environments (e.g., changes in network conditions or resource constraints). For example, when significant changes in the available resource or new users with different characteristics (e.g., users with very high CAQA requirements) are introduced, the model may require retraining or fine-tuning. To address these situations, when performance drops to an unacceptable level (e.g., a 25\% drop in AvgCAQA), a retraining mechanism \cite{ref31} can be employed, wherein the model is updated with training data collected from new scenarios to ensure that it remains effective with evolving conditions.

The pseudocode of LPDRL-F is summarized in Algorithm 1 (see \textbf{Appendix~A}). To further illustrate the proposed algorithm, the detailed designs of the DRL-F algorithm for SGenRA and the RCE algorithm for ComRA are presented in the following subsections.

\subsection{DRL-F for SGenRA Sub-problem}\label{subsec:DRL-F}
DRL-F can be cast into three key elements, i.e., state, action, and LP-guided reward. For SGenRA sub-problem, these elements are defined as follows:

(1)	State: The state $s$ consists of factors such as channel gains ${g_k},\forall k$, which are important for sensing and communication energy allocation. However, the raw channel gain $g_k$ typically spans a wide range, making it difficult for the reinforcement learning algorithm to process directly \cite{ref29}. To address this, a logarithmic transformation is applied to compress the variations in channel gain into a more tractable range. Specifically, ${h_k} = {\rm{lo}}{{\rm{g}}_{10}}\left( {{{{g_k}} \mathord{\left/
 {\vphantom {{{g_k}} L_n}} \right.
 \kern-\nulldelimiterspace} L_n}} \right)$, where ${h_k}$ is the normalized logarithmic channel gain for the $k$-th user and $L_n$ is a normalization parameter based on the ISAC device coverage radius and the path loss model. Also, the minimum requirement on CAQA, i.e., ${\Omega _{k,\min }}$, is taken as part of the state. Thus, we denote the user state by $s = \left[ {{h_1},{\Omega _{1,\min }}, \cdots, {h_K},{\Omega _{K,\min }}} \right]$.

 (2) Action: The action $a = \left\{ {{z_1}, \cdots ,{z_K},{d_1}, \cdots ,{d_K}} \right\}$ represents the generating step and sensing energy allocation for $K$ users, where ${z_k} \in \left[ {{z_{\min }},{z_{\max }}} \right],k \in {\rm{{\cal K}}}$ and ${d_k} \in \left[ {0,1} \right],k \in {\rm{{\cal K}}}$ is the output of the step and sensing branch network, respectively. The original sensing energy action for user $k$ is given by ${E_{k,s}} = {d_k} \cdot {E_{s,\max }}$. However, when ${E_{k,s}} < E_{k,s}^q$, the sensing energy allocation becomes invalid, where $E_{k,s}^q$ is the minimum sensing energy required to ensure that the minimum CAQA constraint ${\Omega _{k{\rm{,min}}}}$ is satisfied when the quality of the generated image is at its maximum value, i.e. ${{{x_k}} \mathord{\left/
 {\vphantom {{{x_k}} {{D_c}}}} \right.
 \kern-\nulldelimiterspace} {{D_c}}} = 1$. According to \eqref{Equ7}, \eqref{Equ8}, and ${n_k} = {{{E_{k,s}}} \mathord{\left/
 {\vphantom {{{E_{k,s}}} {{T_s}{P_s}}}} \right.
 \kern-\nulldelimiterspace} ({{T_s}{P_s}})}$, $E_{k,s}^q$ can be obtained as $ E_{k,s}^q = {T_s}{P_s} \cdot \Upsilon _k^{ - 1}\left( {{\Omega _{k,\min }}/\left( {1 - {\varepsilon _k}\left( {{z_k}} \right)} \right)} \right)$, where $\Upsilon _k^{ - 1}$ is the inverse function of ${\Upsilon _k}$ in terms of ${n_k}$. Thus, an action filter is introduced to map ${d_k}$ into a valid sensing energy range ${A_v} = \left[ {E_{k,s}^q,{E_{s,\max }}} \right]$. After filtering invalid actions, the sensing energy for user $k$ is finally given by
\begin{equation}
    {E_{k,s}} = {d_k} \cdot ({E_{s,\max }} - E_{k,s}^q) + E_{k,s}^q.
\label{Equ10}
\end{equation}

By mapping the original sensing actions to a valid range, the action filter effectively reduces the action space by filtering out invalid actions, which enhances training efficiency.

(3)	LP-guided reward: Given the generating step and sensing energy allocation solutions from the DRL-F algorithm, the communication energy allocation ${{\bf{E}}_c}$ is determined by the RCE algorithm, as outlined in Section~\ref{subsec:RCE}. AvgCAQA is computed based on sensing energy, generating step, and communication energy allocations. Since constraints $\text{C3}$ and $\text{C4}$ are satisfied in action design, and constraints $\text{C1}$, $\text{C2}$, and $\text{C6}$ can be satisfied in the RCE algorithm, the LP-guided reward $r$ is defined to ensure compliance with constraint $\text{C5}$. Specifically, if all users satisfy the individualized minimum CAQA requirement, i.e., ${\Omega _k} \ge {\Omega _{k,\min }}$, the reward is the AvgCAQA. Otherwise, a penalty is introduced based on the number of users violating $\text{C5}$. The reward function is given by
\begin{equation}
r = 
\begin{cases}
\frac{1}{K} \sum\limits_{k=1}^K \Omega_k, 
& \text{if } \Omega_k \ge \Omega_{k,\min},\ \forall k, \\[6pt]
\frac{1}{K} \sum\limits_{k=1}^K \Omega_k - \frac{V}{K},
& \text{otherwise}
\end{cases},
\label{Equ11}
\end{equation}
where $V = \sum\nolimits_k {\left( {{\Omega _k} < {\Omega _{k,\min }}} \right)} $ denotes the number of users violating C5. This reward structure encourages the LPDRL-F algorithm to prioritize solutions that maximizing the AvgCAQA while ensuring the minimum CAQA requirement is met.

A specific DRL algorithm, i.e., the soft actor-critic (SAC) framework, is employed in DRL-F due to its capability to provide improved training stability through entropy regularization \cite{ref32, ref33}, which consists of two critic networks, ${Q_{{\theta _1}}}$ and ${Q_{{\theta _2}}}$, and an actor network ${\pi _\phi }$. At iteration $i$, the algorithm aims to maximize the weighted cumulative discounted reward and policy entropy \cite{ref32}, as follows:
\begin{equation}
    \mathbb{E}\left[ {\sum\limits_{l = i}^\infty  {{\gamma ^{l - i}}[{r_i} - \rho {\rm{log}}{\pi _\phi }({a_i}|{s_i})]} } \right],
\label{Equ12}
\end{equation}
where $\gamma  \in \left( {0,1} \right)$ is the discount factor, and $\rho $ is the temperature parameter. The actor network ${\pi _\phi }$ outputs jointly sensing energy and generating step actions ${a_i}$ based on the agent’s state ${s_i}$, with parameter $\phi $. The critic network, responsible for fitting the agent’s soft Q-function, receives the state ${s_i}$ and action ${a_i}$ as inputs during training. To address the overestimation of the soft Q-function, two critic networks, ${Q_{{\theta _1}}}$ and ${Q_{{\theta _2}}}$, with parameter ${\theta _1}$ and ${\theta _2}$, are employed along with two target critic networks, ${Q_{{{\theta '_1}}}}$ and ${Q_{{{\theta '_2}}}}$, with parameter ${\theta '_1}$ and ${\theta '_2}$. The state-action value function $Q$ is computed based on the soft Bellman equation \cite{ref29}. The critic network parameters are updated as
\begin{equation}
    J({\theta _j}) = \frac{1}{{|{\Psi _b}|}}\sum\limits_{{s_i},{a_i} \in {\Psi _b}} {({y_i} - {Q_{{\theta _j}}}(} {s_i},{a_i}){)^2},j = 1,2,
\label{Equ13}
\end{equation}
where $\left| {{\Psi _b}} \right|$ represents the batch size and ${y_i}$ denotes the target value of critic network, which is given by ${y_i} = {r_i} + \gamma \left( {\mathop {\min }\limits_{j = 1,2} \left( {{Q_{{{\theta '_j}}}}({s_{i + 1}},{a_{i + 1}}) - \rho {\rm{lo}}{{\rm{g}}_{{\pi _\phi }}}({{\tilde a}_{i + 1}}|{s_{i + 1}})} \right)} \right)$, where ${\tilde a_{i + 1}} = {\pi _\phi }\left( { \cdot |{s_{i + 1}}} \right)$ emphasizes that the next action should be resampled from the policy.

The temperature parameter $\rho $ reveals the relative importance of reward versus entropy term, brings randomness to the optimal policy, and can be adaptively optimized according to ${\nabla _\rho }J\left( \rho  \right)$ \cite{ref29}. The actor network parameters are updated according to \cite{ref32}
\begin{equation}
J(\phi ) = \frac{1}{{|{\Psi _b}|}}\sum\limits_{{s_t},{a_t} \in {\Psi _b}} {\left( {\mathop {\min }\limits_{j = 1,2} \left( {{Q_{{\theta _j}}}\left( {{s_i},{a_i}} \right) - \rho {\rm{log}}{\pi _\phi }({a_i}|{s_i})} \right)} \right)} .
\label{Equ14}
\end{equation}

The target critic network is updated according to
\begin{equation}
    {\theta '_j} \leftarrow \varepsilon {\theta _j} + (1 - \varepsilon ){\theta '_j},\;\;\;j = 1,2,
\label{Equ15}
\end{equation}
where $\varepsilon$ is the soft update parameter.

\subsection{RCE Algorithm for ComRA Sub-problem}\label{subsec:RCE}
Given the result of SGenRA sub-problem, the CAQA-AIGC problem is transformed into ComRA sub-problem, optimizing communication energy allocation for each user, given by
\begin{equation}
    \begin{aligned}
    \text{(ComRA):} \quad & \max_{\mathbf{E}_c} \ \frac{1}{K} \sum_{k=1}^{K} \Omega_k \\
    \text{s.t.} \quad 
    & \text{M1: } E_{k,c}^{\min} \le E_{k,c} \le E_{k,c}^{\max}, \quad \forall k, \\
    & \text{M2: } \sum_{k=1}^{K} E_{k,c} \le E_r,
    \end{aligned}
\label{Equ16}
\end{equation}
where constraint M1 is derived from the original C1, C5, and C6 constraints, and constraint M2 is derived from the original C2 constraint in the CAQA-AIGC problem.

\begin{figure*}[ht]
\begin{equation}
    E_{k,c}^{\max } = \min \left( {\underbrace {{P_c}\left( {{T_{\max }} - {T_{k,{\rm{arr}}}} - {T_{k,{\rm{que}}}} - {T_{k,{\rm{gen}}}} - {T_{k,{\rm{wait}}}}} \right)}_{{\text{C1: service time constraint}}},\;\underbrace {\frac{{{P_c}\beta K{D_c}}}{{B{{\log }_2}\left( {1 + {{{g_k}{P_c}} \mathord{\left/
 {\vphantom {{{g_k}{P_c}} {{\delta ^2}}}} \right.
 \kern-\nulldelimiterspace} {{\delta ^2}}}} \right)}}}_{{\text{C6: display constraint}}}} \right).
\label{Equ17}
\end{equation}
\vspace*{-1.0em}
\hrulefill
\vspace*{0.4pt}
\end{figure*}

Specifically, M1 specifies that the minimum communication energy $E_{k,c}^{\min } = \frac{{{\Omega _{k,\min }}{P_c}K\beta {D_c}}}{{{\Theta _k}({E_{k,s}},{z_k}) \cdot B{{\log }_2}(1 + \frac{{{g_k}{P_c}}}{{{\delta ^2}}})}}$ are satisfied for all users, ensuring that each user meets the CAQA requirement (C5). Additionally, the maximum communication energy $E_{k,c}^{\max }$ is determined by the tighter of the service time and display capacity constraints, as defined in \eqref{Equ17}. M2 ensures that the total energy used for communication does not exceed the remaining energy ${E_r} = {E_{{\rm{max}}}} - \mathop \sum \limits_{k = 1}^K {E_{k,s}}$ (total energy minus the total sensing energy). 

ComRA sub-problem aims to optimize the communication energy allocation for each user to maximize the content generation quality (CGQ). To solve this problem, the RCE algorithm is proposed, as detailed in Algorithm~2 (see \textbf{Appendix~B}). The optimality of the RCE algorithm is established in the following theorem.

\vspace{0.6em}
\noindent{\textbf{Theorem 1.}} \textit{The optimal solution to the ComRA sub-problem can be obtained using the RCE algorithm.}

\vspace{0.4em}
\noindent The proof of \textbf{Theorem~1} is provided in \textbf{Appendix~B}.

\subsection{Complexity Analysis}\label{subsec:Complexity Analysis}
For ComRA sub-problem, the computational complexity of standard LP solvers, such as interior-point \cite{ref34} and simplex \cite{ref35} methods, is theoretically bounded by $O\left( {{K^{3.5}}} \right)$ (polynomial) and $O\left( {{2^K}} \right)$ (exponential), respectively. In contrast, the proposed RCE algorithm exploits the inherent structure of ComRA sub-problem, i.e., the linear objective and separable box constraints, to achieve a complexity of $O\left( {K\log K} \right)$ (see \textbf{Appendix~B}). Notably, the complexity reduction does not sacrifice optimality. The allocation strategy provably converges to the global optimum with linear objectives and box constraints (see \textbf{Theorem 1}). This efficiency enables real-time energy allocation in ISAC, where traditional LP tools (e.g., CVXPY, MATLAB’s linprog) become intractable, especially for larger $K$.

When DRL converges, its computational complexity is usually measured by the number of nonlinear transformations in a forward propagation \cite{ref36}, given by $O\left( {\sum\nolimits_{l = 1}^L {{q_l}{q_{l + 1}}} } \right)$ for the proposed DRL-F. Here, $L$ is the number of neural network layers, and ${q_l}$ is the number of neurons in the $l$-th hidden layer. 

For the overall complexity of LPDRL-F, if standard LP solvers, such as simplex methods, are employed to solve ComRA sub-problem, the complexity of simplex methods cannot be ignored, and the overall complexity of LPDRL-F would be $O\left( {{2^K} + \sum\nolimits_{l = 1}^L {{q_l}{q_{l + 1}}} } \right)$. In contrast, the time complexity of our proposed LPDRL-F algorithm can be summarized as $O\left( {K\log K + \sum\nolimits_{l = 1}^L {{q_l}{q_{l + 1}}} } \right)$. Furthermore, the complexity can be further reduced to $O\left( {\sum\nolimits_{l = 1}^L {{q_l}{q_{l + 1}}} } \right)$ in typical immersive VR/XR interaction scenarios. This is because existing DRL typically use 32-256 hidden layer neurons \cite{ref36}. According to 3GPP standards \cite{ref37,ref38}, the number of immersive VR/XR interaction users $K$ is at the order of 10. Thus, $K\log K \ll \sum\nolimits_{l = 1}^L {{q_l}{q_{l + 1}}} $, and the complexity of the RCE algorithm can be neglected.

\section{Performance Evaluation}\label{sec:performance_evaluation}
\begin{table}[ht]
\caption{Simulation Parameters}
\begin{tabular*}{\columnwidth}{@{\extracolsep{\fill}}ll}
\toprule
\textbf{Parameters} & \textbf{Value} \\ \midrule
ISAC device coverage radius & 200 m \\
Number of users $K$ & 10 \\
Total bandwidth $B$ & 100 MHz \\
Maximum service time $T_{\text{max}}$ & 1 s \\
One sensing cycle time $T_s$ & 60 $\mu$s \\
Device display capacity $D_c$ & $512 \times 512$ pixels \\
Maximum total energy $E_{\text{max}}$ & 1 J \\
Maximum sensing energy $E_{s,\text{max}}$ & 0.1 J \\
Bits per pixel $\beta$ & 24 bit \\
Minimum CAQA $\Omega_{k,\text{min}}$ & $\sim U[0.4, 0.45]$ \\
Minimum generation step $z_{\min}$ & 5 \\
Maximum generation step $z_{\max}$ & 10 \\
Average FLOPs per step $\chi$ & 0.2 TeraFLOPS/step \\
Computation capacity of server $f$ & 20 TeraFLOPS \\
Scaling factor $\epsilon^{\rm{fwd}}$ & 0.03 \\
Attenuation factor $\mu$ & 0.2 \\
Sensing power $P_{s}$ & 1 W \\
Communication power $P_{c}$ & 1.5 W \\
\bottomrule
\end{tabular*}
\label{table:parameters}
\end{table}
Simulation parameters of the ISAC-based AIGC environment and LPDRL-F are listed in Table~\ref{table:parameters} and Table~\ref{tab:2}, respectively \cite{ref17,ref18,ref20,ref21,ref39,ref40}. Large-scale path loss is given by $128.1 + 37.6{\rm{lo}}{{\rm{g}}_{10}}\left( {d/1000} \right)$, where $d$ is the distance between the ISAC device and the user in meters \cite{ref31}. ${\delta ^2}$ is -174 dBm/Hz. The simulation consists of 6000 training episodes, each with 50 iterations, and 10000 testing episodes with 10 iterations. To fully evaluate the algorithm with diverse scenarios, user positions are randomly reset at the beginning of each episode, and users' task priority level and CAQA requirements are reset at each time slot. Unless otherwise stated, all simulations are conducted with the above settings.

\begin{table}[htbp]
\centering
\caption{\label{tab:2}Main Hyperparameters of LPDRL-F}
\begin{tabular*}{\columnwidth}{ll}
\toprule
\textbf{Component} & \textbf{Configuration} \\
\midrule
\multicolumn{2}{l}{\textit{Actor Network (Hybrid Action)}} \\
Input dimension & $2K$ \\
Shared Hidden layers & [256, 128] neurons \\
Sensing branch & [64] neurons \\
Step branch & $K \times |z_{\max}-z_{\min}+1|$ logits \\
Output dimension & $2K$ (sensing + step) \\
\midrule
\multicolumn{2}{l}{\textit{Critic Network}} \\
Input dimension & $4K$ \\
Hidden layers & [256, 128] neurons \\
Output dimension & 1 \\
\midrule

\multicolumn{2}{l}{\textit{Training Hyperparameters}} \\
Actor learning rate & $1 \times 10^{-4}$\\
Critic learning rate & $1 \times 10^{-3}$ \\
Replay buffer size $\Psi$ & 50000 \\
Minimum training size $\Psi_{\min}$ & 5000 \\
Batch size $\left| {{\Psi _b}} \right|$ & 256 \\
Soft update parameter $\varepsilon$ & 0.005 \\
Initial temperature parameter & 0.01 \\
\bottomrule
\end{tabular*}
\end{table}

To evaluate the performance of the proposed LPDRL-F, the following algorithms are employed as baselines:
\begin{itemize}
\item Joint generative diffusion model with action filter (\textbf{JGDM-F}): JGDM-F uses a generative diffusion model (GDM) \cite{ref6,ref14} with an action filter to generate sensing, generating, and communication resource allocation schemes. JGDM-F is the state-of-the-art algorithm.

\item Joint DRL with action filter (\textbf{JDRL-F}): Sensing, generating, and communication resource allocation are solved with DRL (i.e., using soft actor-critic (SAC) \cite{ref13,ref41}), combined with an action filter. Comparing JDRL-F with LPDRL-F can show the advantage of the LP-based approach in solving ComRA sub-problem.

\item LPDRL-F without action filter (\textbf{LPDRL}): Sensing, generating, and communication resource allocation are solved with LPDRL-F without action filter. Comparing LPDRL with LPDRL-F can show the advantage of the action filter approach in solving SGenRA sub-problem. 

\item DRL-F with Uniform communication (\textbf{UDRL-F}): Sensing and generating resource allocation (i.e., optimizing sensing energy and generating step) are solved using DRL-F, and the remaining energy is evenly allocated to all users for communication after meeting the minimum CAQA requirement.

\item CGQ-Only oriented algorithm with fixed sensing energy and generating step (\textbf{CGQ-FSG-$\boldsymbol{\alpha}$}): Consider a CGQ-only oriented scheme. Since perfect content accuracy is assumed, sensing energy is fixed as a proportion $\alpha  \in \left( {0,1} \right)$ of the per-user energy (i.e., ${{{E_{\max }}} \mathord{\left/
 {\vphantom {{{E_{\max }}} K}} \right.
 \kern-\nulldelimiterspace} K}$) \cite{ref42}, and the generating step is fixed to $\left\lfloor {{{\left( {{z_{\min }} + {z_{\max }}} \right)} \mathord{\left/
 {\vphantom {{\left( {{z_{\min }} + {z_{\max }}} \right)} 2}} \right.
 \kern-\nulldelimiterspace} 2}} \right\rfloor $. Then, RCE algorithm is employed to obtain the optimal CGQ-only oriented communication energy allocation. Comparing CGQ-FSG-$\alpha$ with CAQA-AIGC can show the importance of considering content accuracy in ISAC-AIGC networks.

\item Sensing accuracy and quality aware with fixed generating step (\textbf{SAQA-FG}): Consider the impact of sensing accuracy and communication quality while assuming perfect generation without error. i.e., AEG=0 \cite{ref20}. Sensing energy allocation is solved using DRL-F with an action filter, and ComRA is optimized by LP. Since AEG is not considered, the generating steps are fixed to a given value. Comparing SAQA-FG with CAQA-AIGC can show the importance of considering AEG in ISAC-AIGC networks.
\end{itemize}
\begin{figure}[!t]
    \centering 
    \includegraphics[width=3.3in]{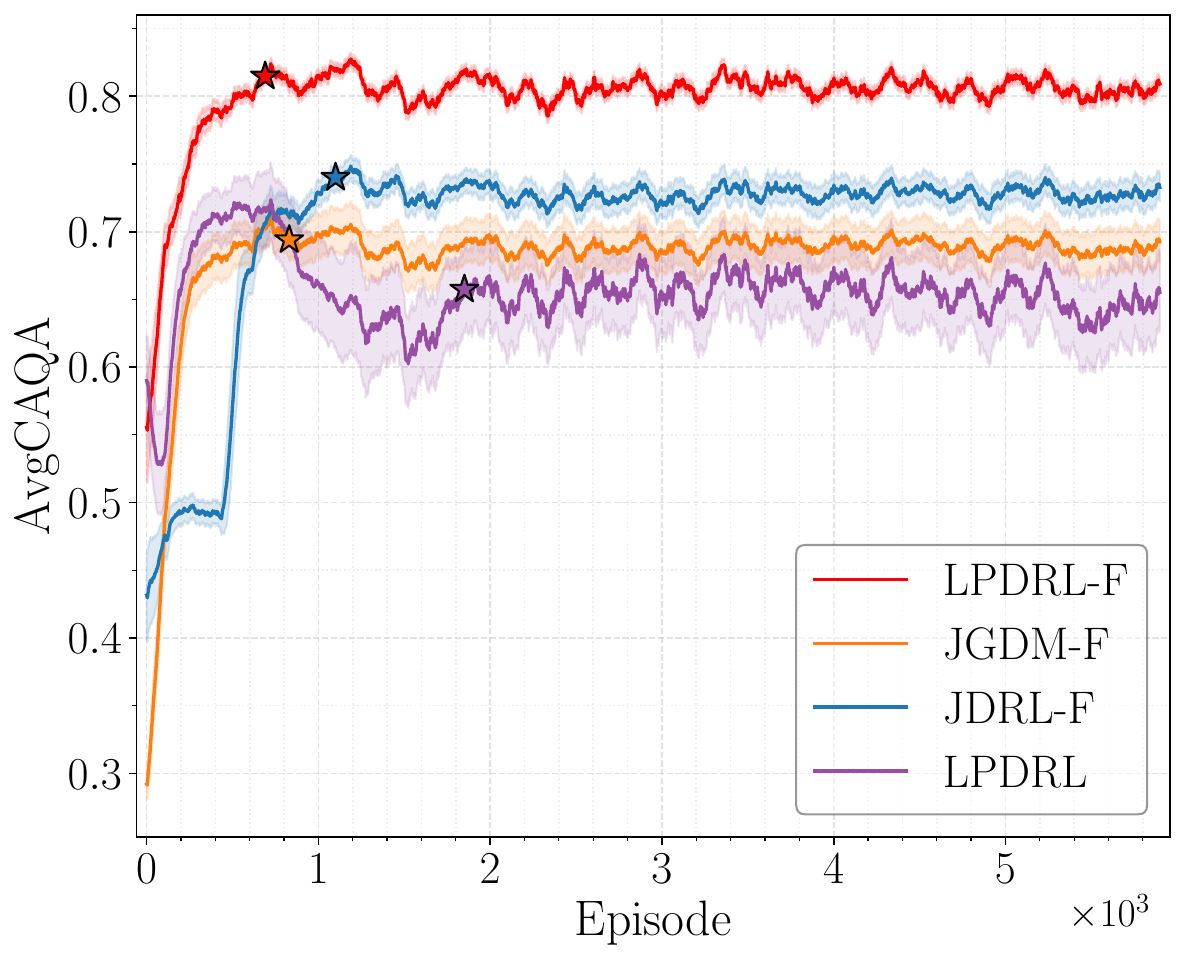}
    \caption{Convergence performance.}
    \label{Fig5}
\end{figure}

First, Fig. \ref{Fig5} illustrates the convergence performance of various algorithms, where the lines represent the average AvgCAQA and the shaded regions indicate the standard deviation of AvgCAQA. A smaller shaded area corresponds to better training stability and lower performance variation. It can be seen that among all investigated algorithms, the proposed LPDRL-F achieves the best AvgCAQA performance, convergence speed, and training stability. This is because in LPDRL-F, LP guidance not only reduces the solution space for communication energy allocation but also obtains the optimal allocation. In contrast, JDRL-F must use the DRL-F algorithm to simultaneously search for sensing, generating, and communication resource allocation solutions, which may not be optimal. Similarly, JGDM-F uses GDM to simultaneously search for sensing, generating, and communication resource allocation solutions. Due to GDM’s powerful generative capabilities, it quickly converges within the first 1000 episodes, which is faster than JDRL-F and LPDRL. However, it ultimately achieves lower AvgCAQA performance than both JDRL-F and LPDRL-F. This is because on one hand, the diffusion-based GDM rapidly learns to generate feasible solutions, on the other hand, it tends to converge prematurely to suboptimal policies without sufficient fine-grained policy exploration. Moreover, it can be seen that the training stability (i.e., shaded region) of LPDRL without the action filter performs the worst. This is because the action filter narrows the solution space for sensing energy allocation, facilitating efficient exploration and the discovery of valid solutions. Without the action filter, LPDRL is stuck with sub-optimal solutions. Overall, Fig. \ref{Fig5} validates the effectiveness of both LP guidance and the action filter in LPDRL-F, which outperforms state-of-the-art methods such as JGDM that jointly optimize the three-dimensional action space. 

\begin{figure}[!t]
    \centering
    \includegraphics[width=3.3in]{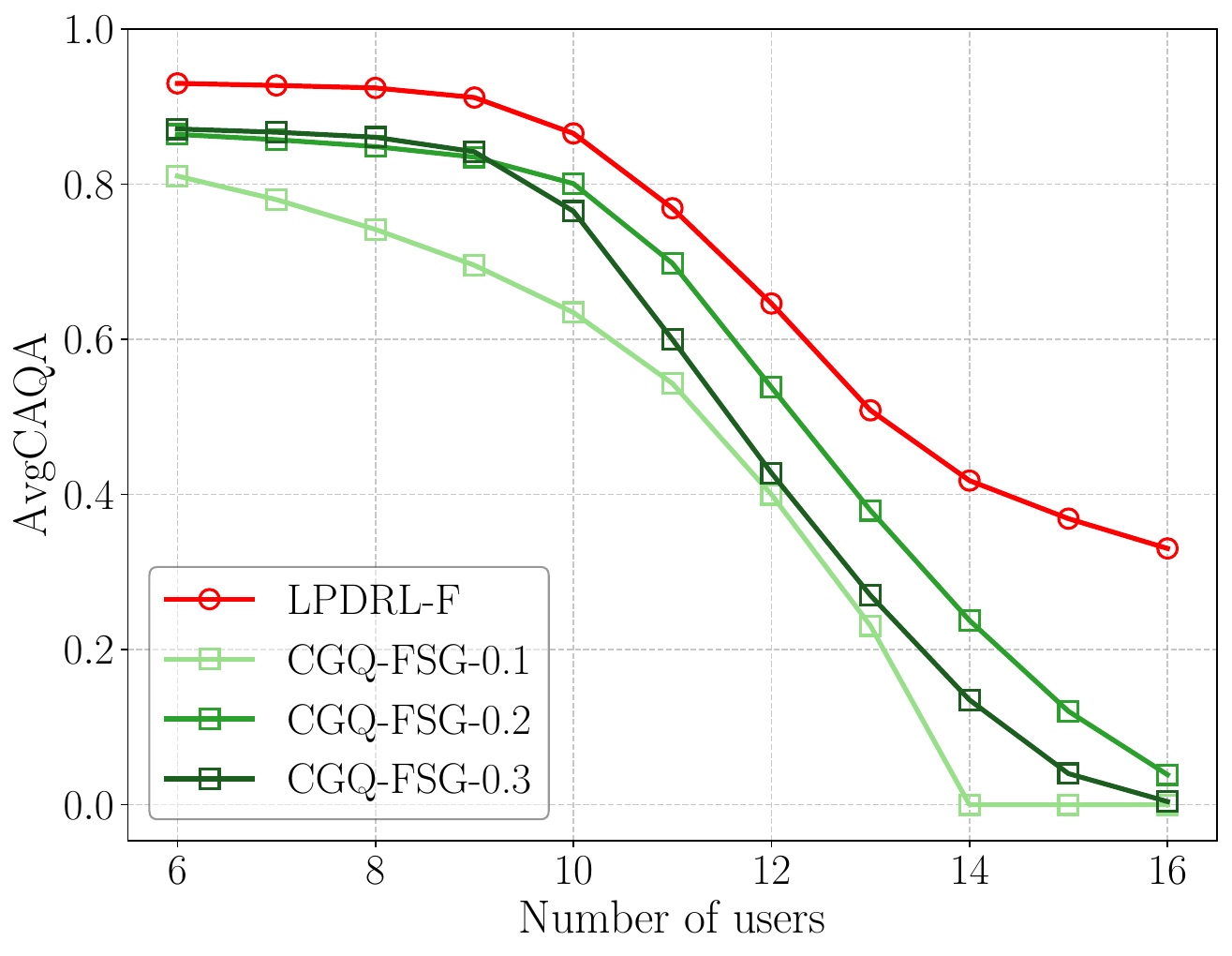}
    \caption{Impact of the number of users on AvgCAQA.}
    \label{Fig6}
\end{figure}

\begin{figure}[!t]
    \centering    
    \includegraphics[width=3.3in]{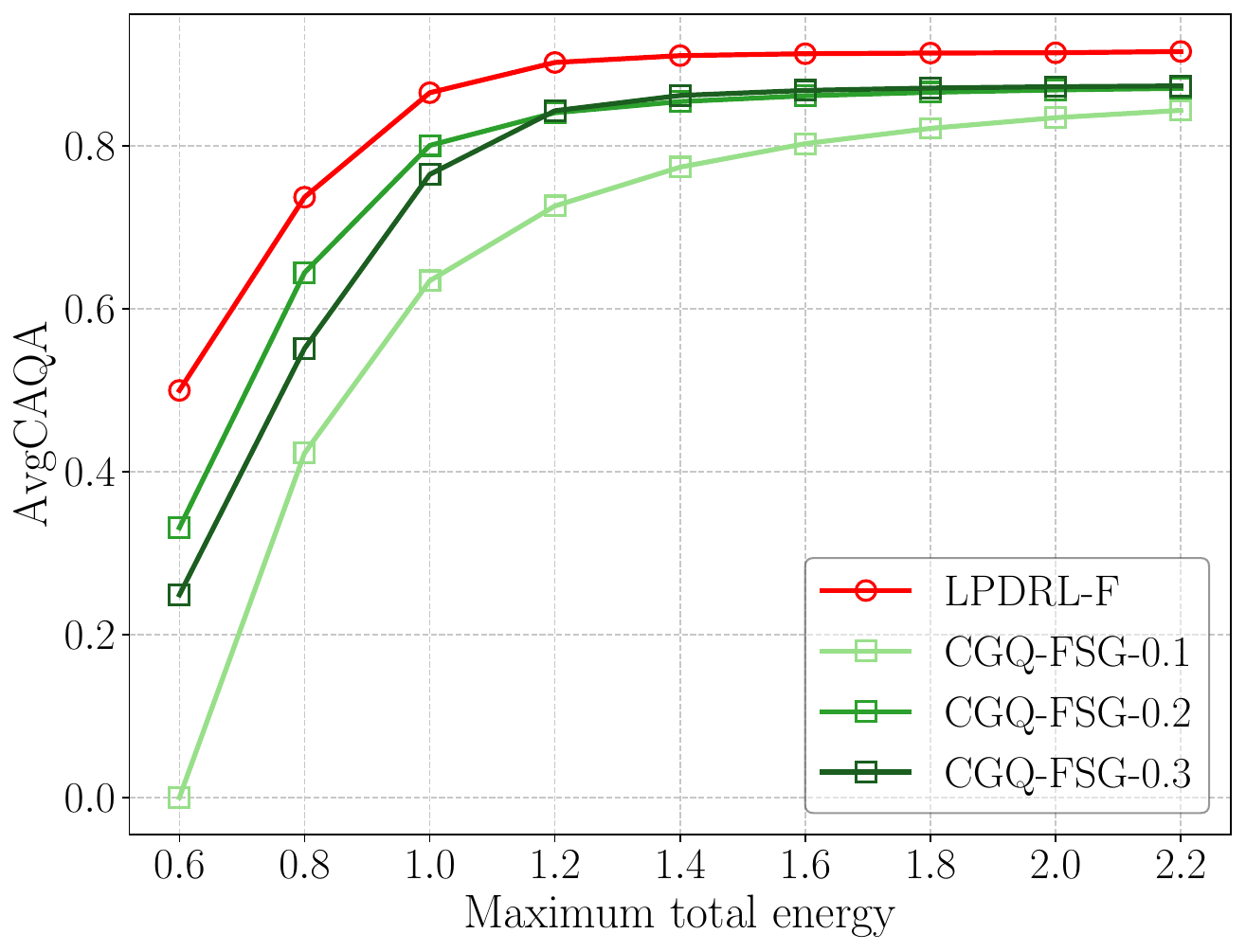}
    \caption{Impact of the maximum total energy on AvgCAQA.}
    \label{Fig7}
\end{figure}

\begin{figure}[!t]
    \centering
    \includegraphics[width=3.3in]{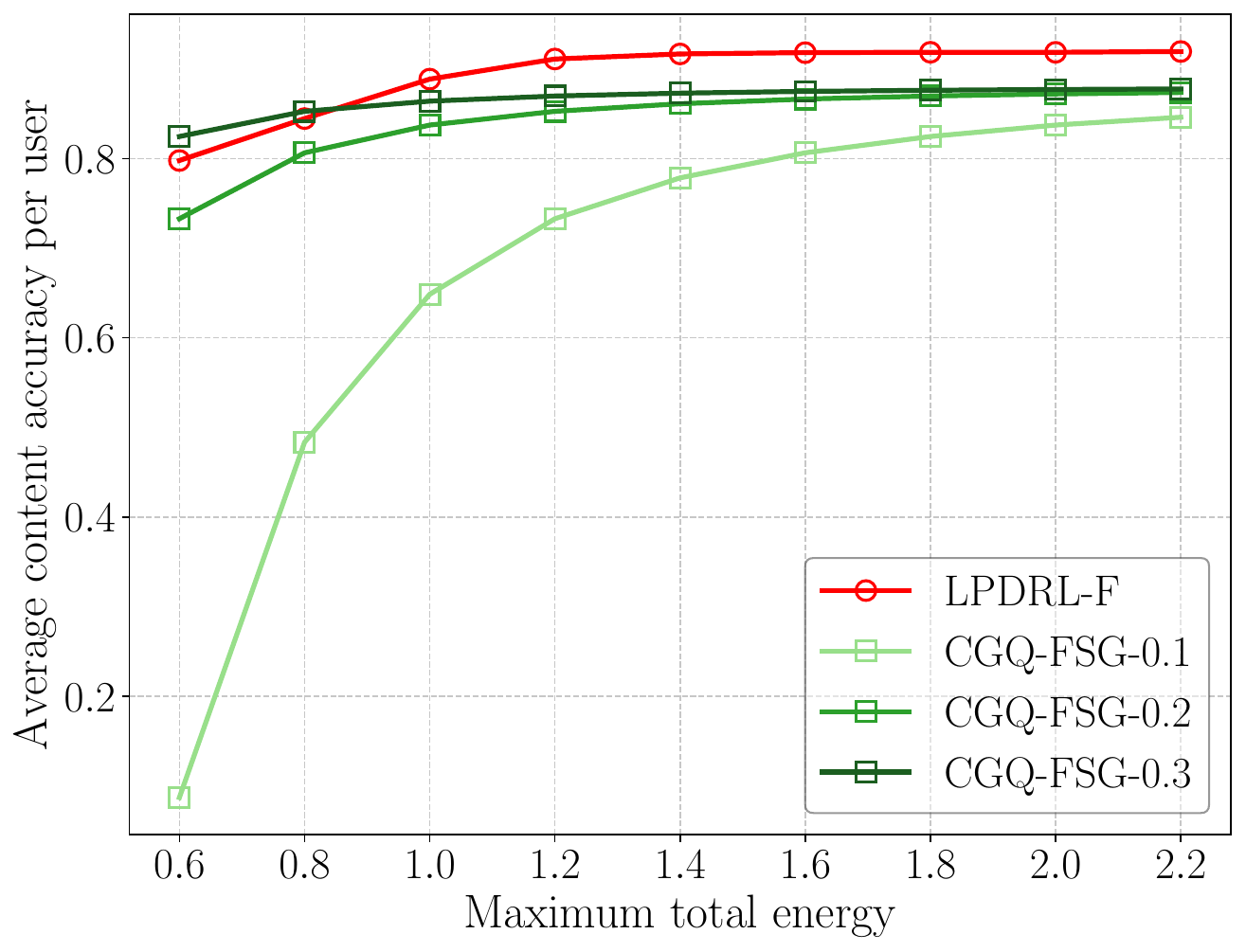}
    \caption{Impact of the maximum total energy on content accuracy.}
    \label{Fig8}
\end{figure}

\begin{figure}[!t]
    \centering
    \includegraphics[width=3.3in]{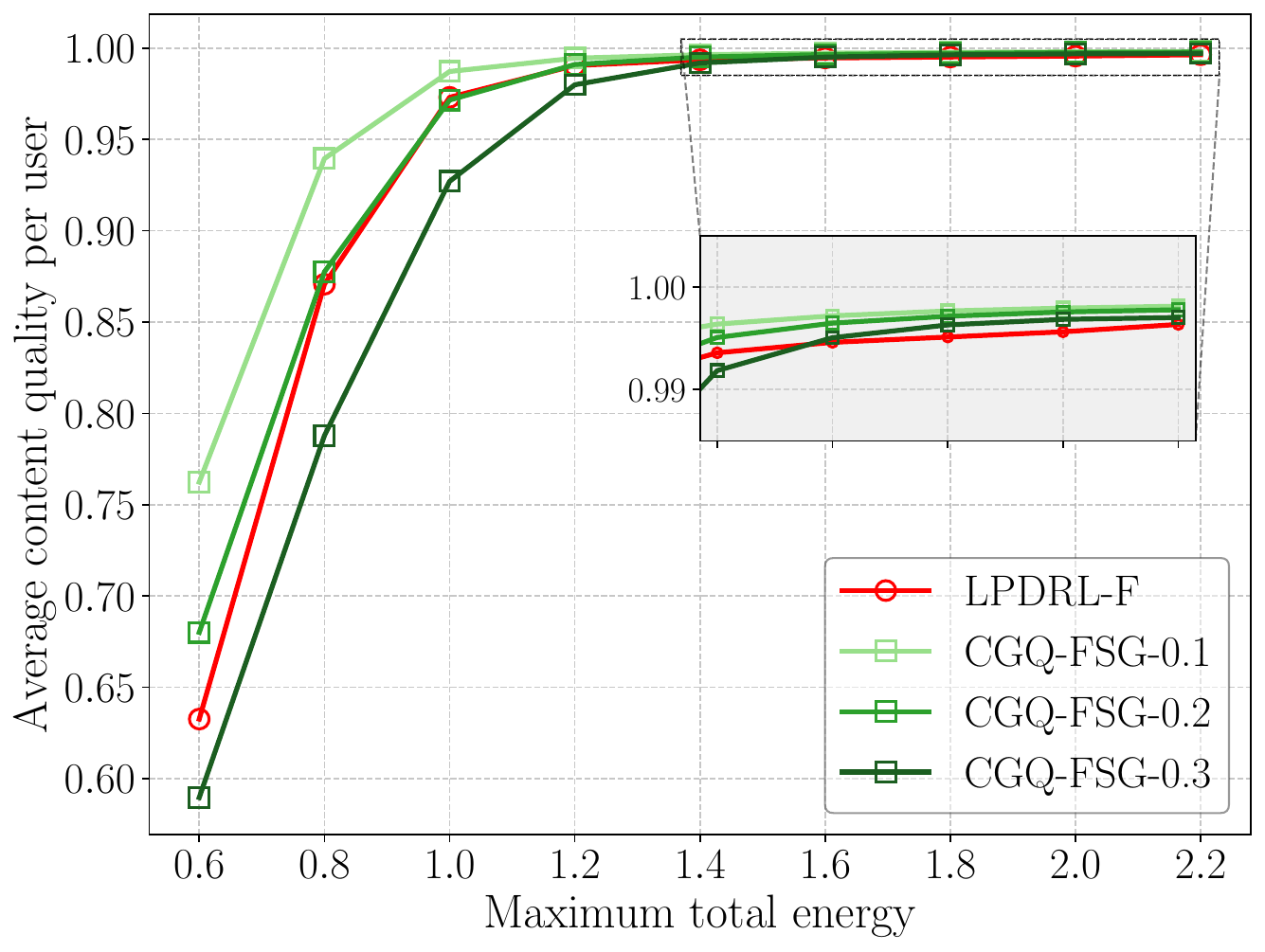}
    \caption{Impact of the maximum total energy on content quality.}
    \label{Fig9}
\end{figure}

\begin{figure}[!ht]
    \centering
    \includegraphics[width=3.3in]{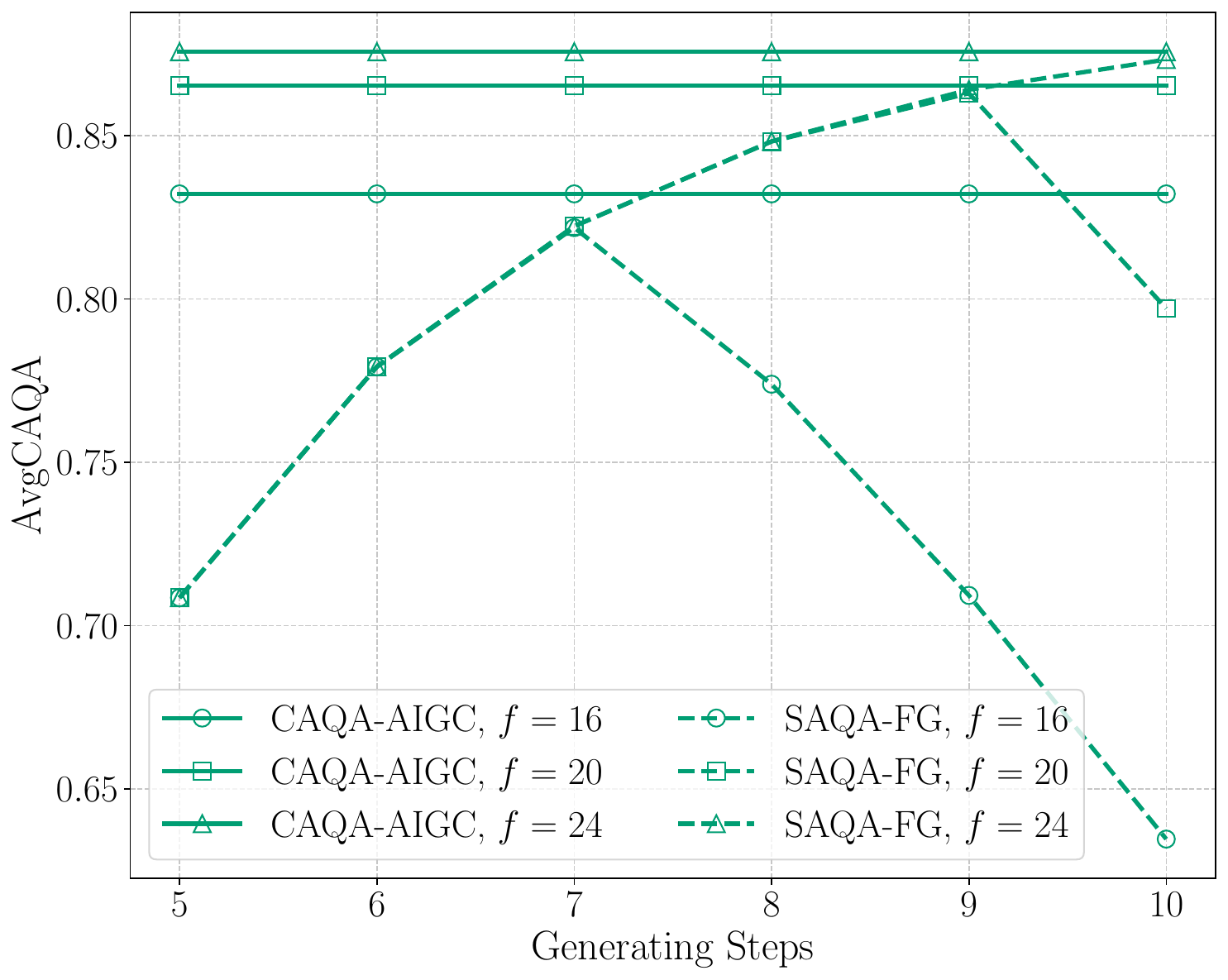}
    \caption{AvgCAQA versus generating step $z$ with different server computation capacities $f$.}
    \label{Fig10}
\end{figure}

\begin{figure}[!ht]
    \centering
    \includegraphics[width=3.5in]{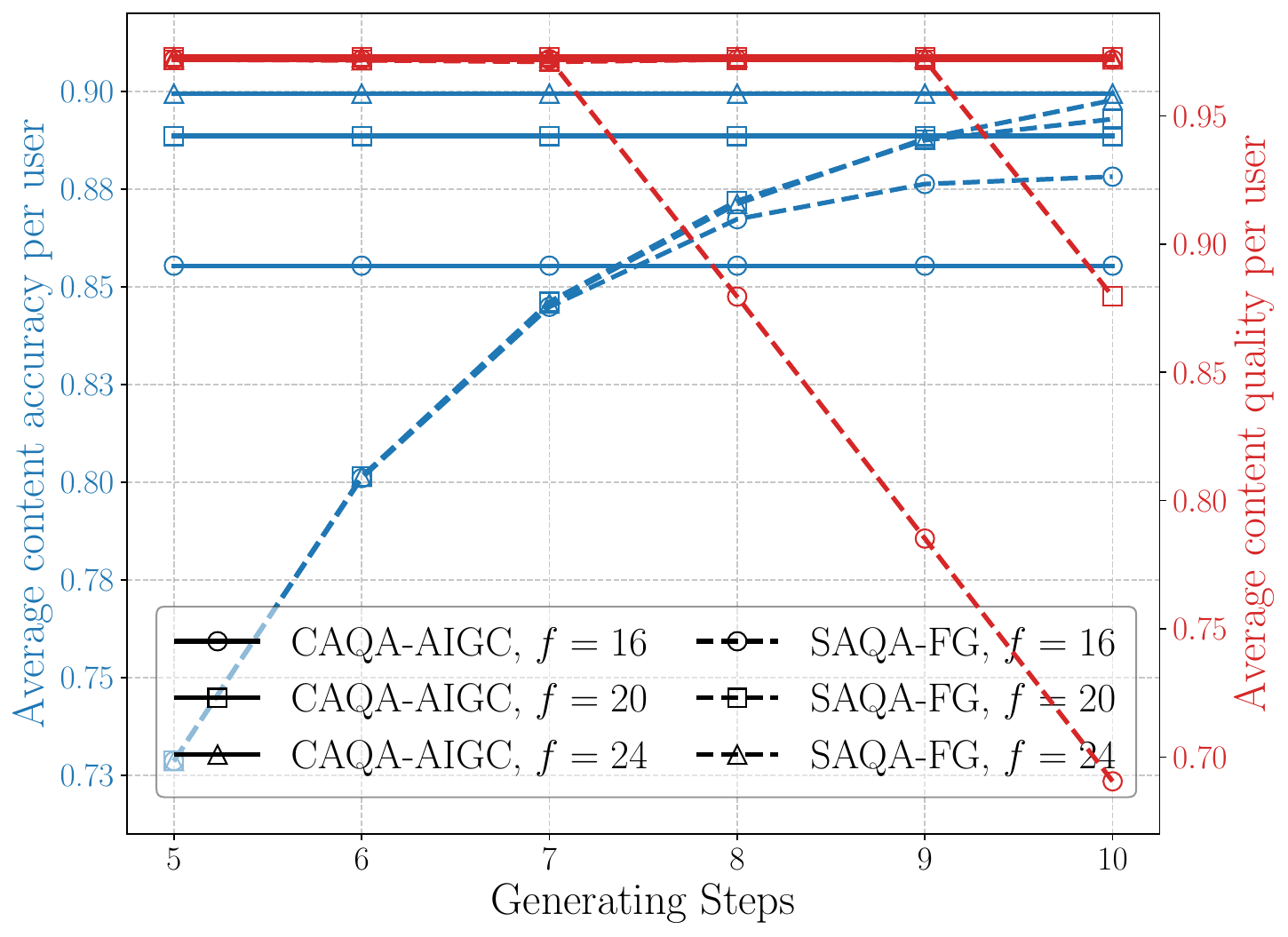}
    \caption{Accuracy and quality versus generating step $z$ with different server computation capacities $f$.}
    \label{Fig11}
\end{figure}

It can be concluded that the proposed LPDRL-F achieves better convergence and higher AvgCAQA than the generative diffusion model (GDM) and DRL baselines. Therefore, the superiority of the proposed LPDRL-F over SOTA algorithms such as GDM and RL is verified. Then, in Figs. \ref{Fig6}--\ref{Fig11}, we tried to verify the effectiveness and importance of introducing the content accuracy in the service assessment metric, i.e., the proposed content accuracy and quality aware service assessment metric (CAQA). Therefore, these figures compare the proposed CAQA-AIGC scheme (with the proposed LPDRL-F algorithm) with existing schemes that consider only content generation quality (CGQ) (with a CGQ-FSG-$\alpha$ algorithm \cite{ref42}) or both sensing accuracy and content generation quality (SAQA) (with a SAQA-FG algorithm \cite{ref20}).

To evaluate the effectiveness of the proposed CAQA-AIGC and show the impact of content accuracy on the system performance, CAQA-AIGC with LPDRL-F and CGQ-oriented baselines are compared in Fig. \ref{Fig6}, where the AvgCAQA performance is shown as a function of the number of users $K$. As $K$ increases, AvgCAQA decreases for all schemes due to the reduced resources (energy, generating capacity, and communication bandwidth) available to each user. CAQA-AIGC with LPDRL-F consistently outperforms the others and shows a slower degradation trend compared to CGQ-FSG-$\alpha$. When the number of users is small, CGQ-FSG with larger $\alpha$ performs better, considering $\alpha$=0.1, 0.2, and 0.3, as sufficient energy is available to ensure high generating accuracy (close to its assumption of perfect accuracy), while maintaining adequate communication quality. As $K$ increases, both large (0.3) and small (0.1) values of $\alpha$ result in worse performance compared to that with $\alpha$=0.2. This is because when $\alpha$ is too small, insufficient energy is allocated for sensing to provide high-accuracy input to generation (its assumption of perfect accuracy totally fails), while excessive $\alpha$ leads to insufficient residual energy for communication to ensure content quality (content accuracy and quality are coupled in a resource-limited real system). Moreover, as $K$ increases, the performance of CGQ-FSG degrades rapidly. This is because CGQ-FSG always allocates fixed sensing energy and generating step regardless of the number of users, leading to insufficient residual energy and time for communication. When $K \ge 14$, CGQ-FSG-0.1 fails to meet the CAQA requirements of any user, leading to an AvgCAQA of zero. Different to CGQ-FSG, CAQA-AIGC considers imperfect content accuracy and jointly optimizes the sensing-generation-communication resource allocation to achieve good content accuracy and quality. Thus, CAQA-AIGC always outperforms CGQ-FSG-$\alpha$ in real systems with imperfect content accuracy. It can be seen that content accuracy has significant impact on the system performance, and it is important to consider content accuracy in resource allocation schemes.


\begin{figure}[!t]
    \centering
    \includegraphics[width=3.3in]{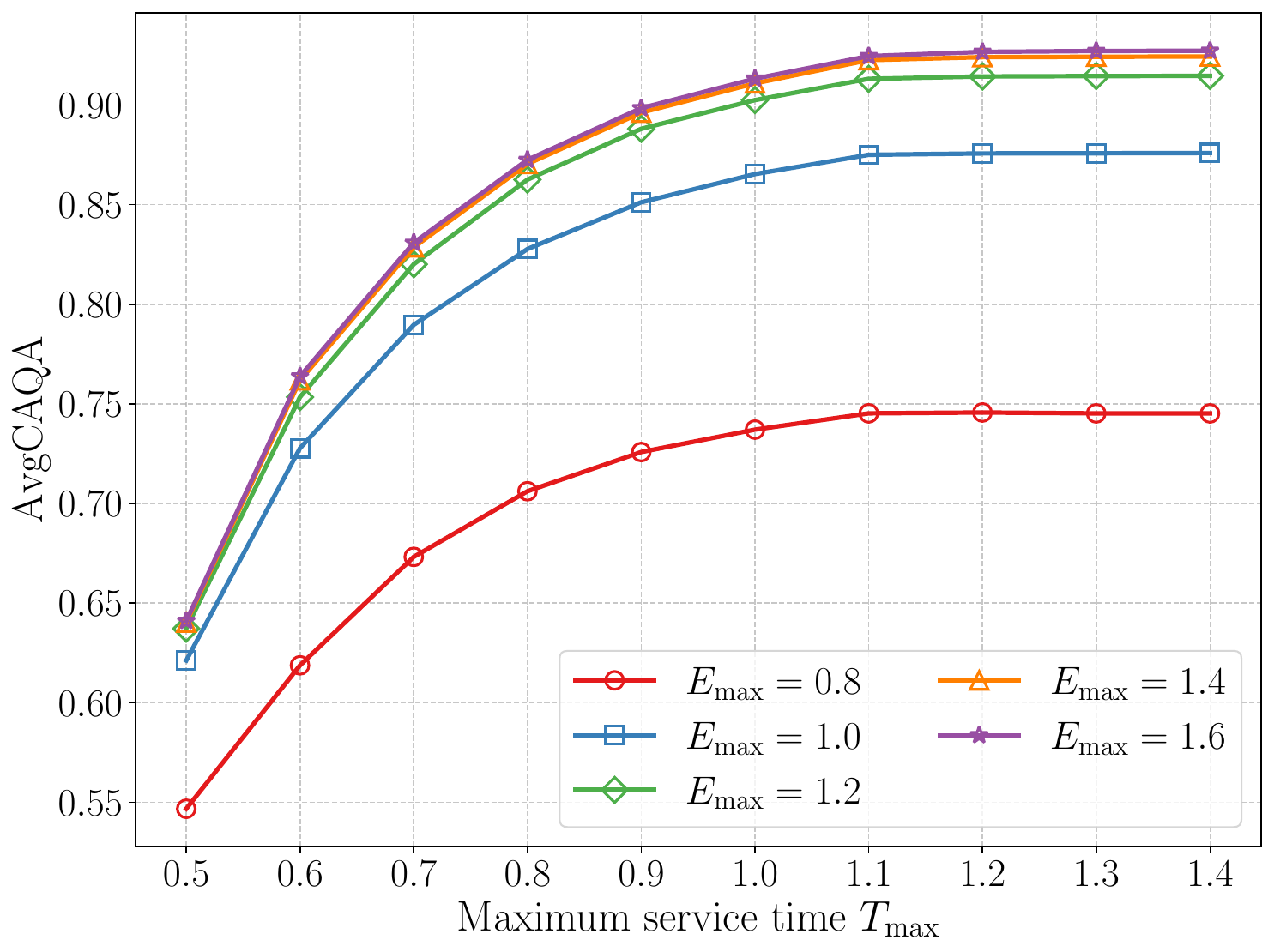}
    \caption{AvgCAQA versus maximum service time ${T_{{\rm{max}}}}$ with different total energies.}
    \label{Fig12}
\end{figure}

Then, Fig. \ref{Fig7} shows the AvgCAQA performance as a function of the maximum total energy ${E_{\max }}$. It can be seen that for all investigated algorithms, AvgCAQA improves as ${E_{\max }}$ increases, due to more available energy for both sensing and communication. As ${E_{\max }}$ continues to increase, the gain in AvgCAQA gradually saturates. This is because AvgCAQA is jointly constrained by both maximum energy and service time (as shown in C1--C3 in \eqref{Equ9}), which are coupled with each other. Note that with a fixed maximum service time, the energy that can be utilized is inherently capped. Thus, as the energy continues to increase and reaches this usable limit, further increasing the energy no longer yields noticeable improvement. As shown in Fig. \ref{Fig12}, further AvgCAQA gains may be achieved by increasing the maximum service time. The proposed CAQA-AIGC with LPDRL-F consistently achieves the highest AvgCAQA. Similar to Fig. \ref{Fig6}, this is because with the assumption of perfect content accuracy, CGQ-FSG-$\alpha$ is designed to ensure a fixed sensing energy proportion and then allocate the remaining energy for communication based on a CGQ-oriented strategy. In a practical system with imperfect content accuracy, when the total energy is small, allocating a larger proportion of sensing energy can achieve higher content accuracy (see Fig. \ref{Fig8}). However, it also results in lower content quality since less resource is available for communications (see Fig. \ref{Fig9}). Compared to CGQ-FSG-$\alpha$, CAQA-AIGC with LPDRL-F can achieve a better balance between the content accuracy and quality and outperforms CGQ-FSG-$\alpha$ when ${E_{\max }}$ is small. As ${E_{\max }}$ increases, CGQ-FSG-$\alpha$ can allocate sufficient energy for communication and enhance content quality (see Fig. \ref{Fig9}). However, since a fixed sensing energy is employed by CGQ-FSG-$\alpha$, it cannot fully exploit the advantage of high total energy to further enhance content accuracy. Different to CGQ-FSG-$\alpha$, CAQA-AIGC is aware of the content accuracy in a practical system. It can optimize the sensing-generating-communication resource allocation to fully exploit the high total energy. As shown in Fig. \ref{Fig8} and Fig. \ref{Fig9}, when ${E_{\max }}$ is large, CAQA-AIGC achieves a notably better content accuracy with a slightly worse content quality. Overall, the AvgCAQA performance of CAQA-AIGC is always better than that of CGQ-FSG-$\alpha$.

To evaluate the impact of AEG (one of the sources affecting content accuracy) and the effectiveness of adaptive generating step optimization, Fig. \ref{Fig10} shows AvgCAQA as a function of the generating steps with different computing capacities. SAQA-FG uses a fixed number of generating steps while optimizing sensing and communication energy. As the generating steps increase, its AvgCAQA first improves due to higher generation accuracy (see Fig. \ref{Fig11}), then degrades because longer generation time reduces the time available for communication, which lowers content quality (see Fig. \ref{Fig11}). It should be noted that with CAQA-AIGC, instead of using fixed generating steps, the generating steps are jointly OPTIMIZED with sensing and communication resources, yielding consistently high AvgCAQA with various computing capacities. Moreover, the optimal generating step in SAQA-FG increases with computing capacity $f$. For $f$ = 16, the optimal step is 7, whereas for $f$ = 20, it increases to 9. This is because a larger $f$ can support a larger generating step without impacting the communication time (content quality), thereby mitigating the impact of AEG and allowing the assumption of perfect generating accuracy to approximately hold. It can be seen that the system performance is sensitive to the generating step, which is a key factor in balancing content accuracy and quality (see Fig. \ref{Fig11}). Thus its optimization is crucial for maximizing AvgCAQA.

To evaluate the impact of time and energy constraints on the performance of CAQA-AIGC, Fig. \ref{Fig12} shows AvgCAQA as a function of the maximum service time ${T_{{\rm{max}}}}$ with different maximum total energy ${E_{\max }}$. As ${T_{{\rm{max}}}}$ increases, the performance gap between different ${E_{\max }}$ widens. This is because, when ${T_{{\rm{max}}}}$ is low, CAQA is primarily constrained by time availability, and increasing ${E_{\max }}$ has little effect on AvgCAQA due to the lack of sufficient time. However, as ${T_{{\rm{max}}}}$ increases, a longer service time becomes available, and more energy leads to a more noticeable improvement in AvgCAQA.

\begin{figure}[!t]
    \centering
    \includegraphics[width=3.3in]{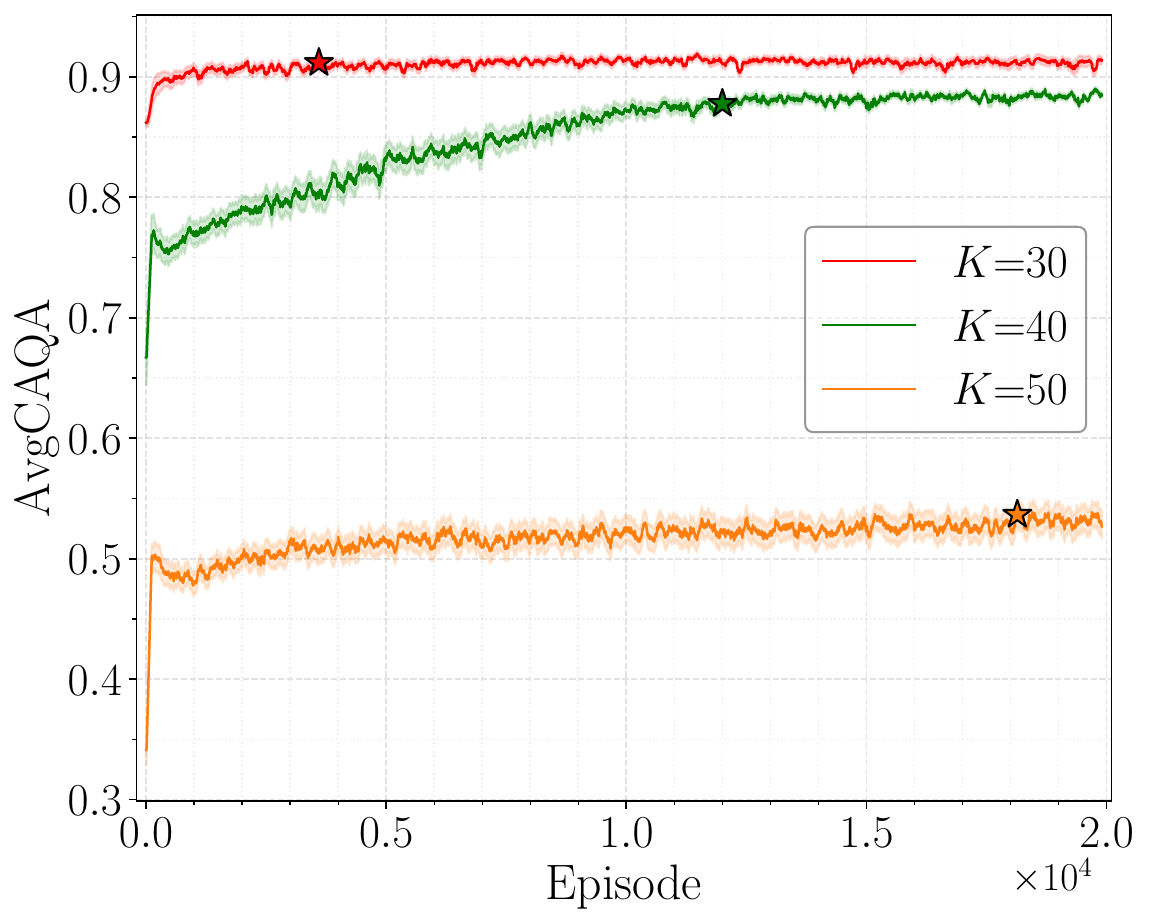}
    \caption{Convergence of LPDRL-F with different numbers of users.}
    \label{Fig13}
\end{figure}

\begin{figure}[!t]
    \centering
    \includegraphics[width=3.3in]{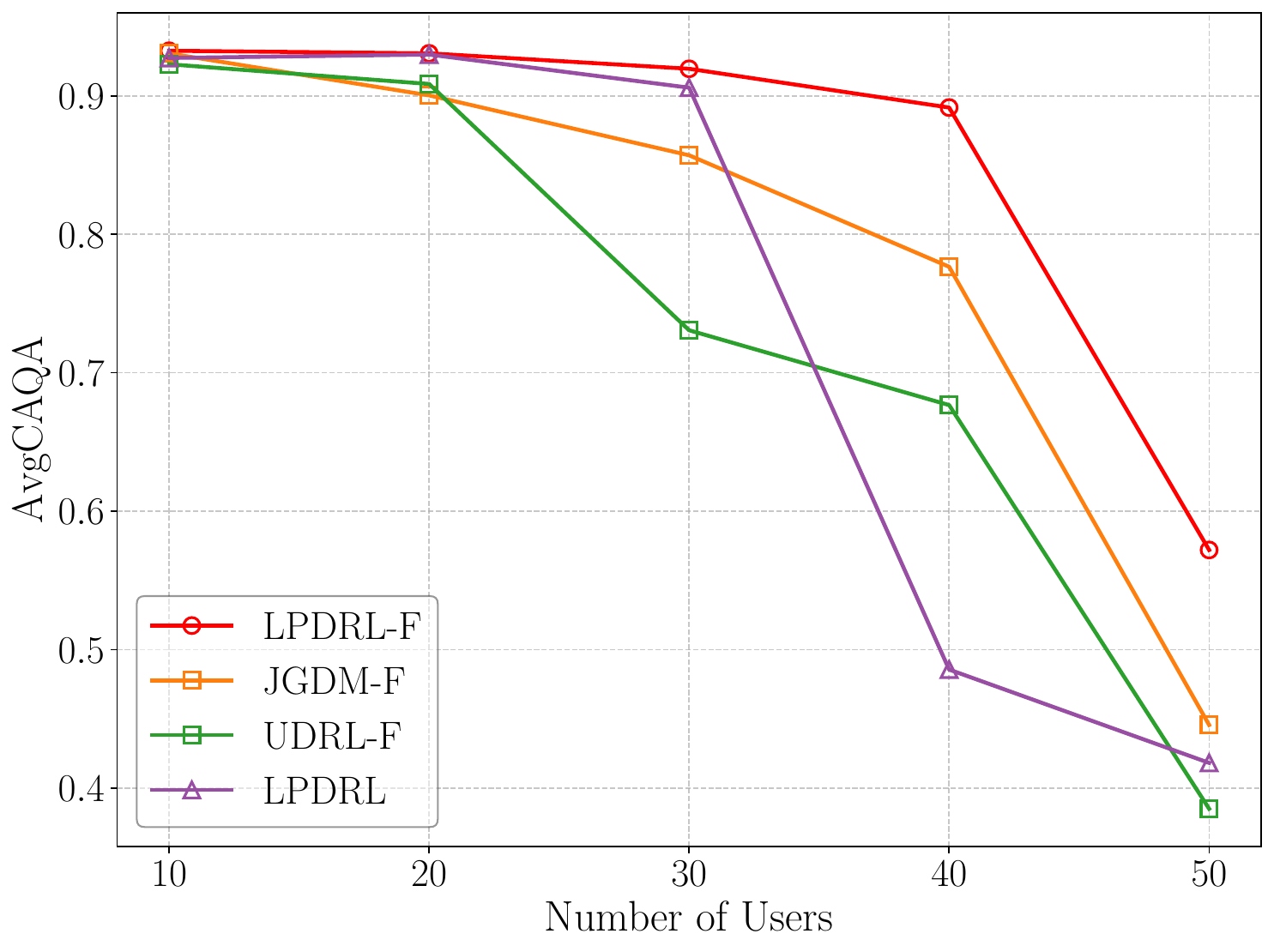}
    \caption{AvgCAQA versus the number of users.}
    \label{Fig14}
\end{figure}

Note that according to 3GPP standard \cite{ref37,ref38}, typically around ten VR users are assumed. When larger user scales are considered, more resources are needed. So  Fig. \ref{Fig13} and Fig. \ref{Fig14} are obtained with the settings of ${E_{\max }} = 4.0$, ${T_{\max }} = 4.0$, $B = 500{\mathop{\rm MHz}\nolimits}$, and $2 \times {10^4}$ training episodes. Fig. \ref{Fig13} shows the convergence performance of the proposed LPDRL-F with larger user scales, where the lines represent the average AvgCAQA and the shaded regions indicate the standard deviation of AvgCAQA. A smaller shaded area stands for better training stability and lower performance variation. As the number of episodes increases, the standard deviation decreases. This is because the policy of LPDRL-F becomes increasingly stable as interactions with the environment increase. As the number of users $K$ increases, the shaded area increases, and so do the training episodes (training time) required to reach convergence. This is because a larger $K$ introduces stronger inter-user coupling in the sensing, generation, and communication resource allocation. The dimension of decision variables is $2K$, resulting in a more complex state-action mapping and increased learning difficulty, which leads to higher performance variability in the early stages and slows the convergence.

Fig. \ref{Fig14} shows the performance of the proposed LPDRL-F and the comparing algorithms as a function of the number of users $K$. It can be seen that the performance of all concerned algorithms decreases as $K$ increases, because resource contention intensifies. The proposed LPDRL-F achieves the best AvgCAQA performance. The AvgCAQA of LPDRL without the action filter drops sharply, indicating that the filter is essential for removing infeasible actions, especially at large $K$. This is because, as $K$ increases, resource contention becomes more intense and the proportion of infeasible actions grows, making the action filter increasingly effective. Meanwhile, the AvgCAQA of UDRL-F, which allocates communication resources uniformly, is worse than that of LPDRL-F, which yields optimal communication resource allocations with the help of LP guidance. Overall, the results show that LP guidance and the action filter are both necessary to sustain high performance. When $K$ is small ($K \le 30$), LPDRL outperforms UDRL-F because resources are relatively sufficient, and the LP module can exploit communication resources more effectively than uniform allocation, making the LP more effective.

\section{Conclusion and Discussion}\label{sec:conclusion}
Considering the imperfect content accuracy in ISAC-based AIGC networks caused by inaccurate sensing input and the inherent AEG during the generating process, this paper proposed a content accuracy and quality aware service metric CAQA. Based on CAQA, a joint sensing, generating, and communication resource allocation problem (CAQA-AIGC) was formulated, aiming to maximize average CAQA (AvgCAQA). To solve this problem with low complexity, the LPDRL-F algorithm was proposed. Compared with existing DRL-based and GDM-based algorithms, LPDRL-F achieves over 10\% improvement in AvgCAQA by operating in a smaller and more targeted solution space. Moreover, compared to the CGQ-only oriented scheme, the proposed CAQA-AIGC with LPDRL can improve AvgCAQA by over 50\%. This study demonstrates that in practical ISAC-based AIGC networks with imperfect content accuracy caused by inaccurate sensing and AEG, CAQA can serve as an effective performance metric to guide multi-domain resource co-optimization (sensing, computing, and communication), thereby enhancing the overall service quality of generated content.

The overhead of the LPDRL-F agent’s state collection and resource-decision conveying is critical \cite{cui2025overview,tian2023efficient}. First, the entire LPDRL-F agent, including the DRL-F module for SGenRA and the RCE module for ComRA, is executed on the edge ISAC device. Considering the overhead of state collection for the agent, since the agent only requires each user’s channel gain and a single scalar-valued CAQA requirement, this information can be reported by each user to the agent with an uplink transmission of tens of bytes, resulting in low communication overhead \cite{liu2024multidimensional}. For resource decision conveying, since the sensing and communication allocations are executed at the edge ISAC device, no communication overhead is incurred. The AIGC generation-step decisions are executed at the AIGC server. This requires transmitting only $K$ scalar values (one per user) over the backhaul between the ISAC device and the AIGC server, so this overhead is negligible compared to the transmission of raw AIGC data \cite{sun2025aerial,hou2024distributed}. Since this paper focuses on resource allocation, a subjective validation is beyond the scope of this paper. Such validation may be incorporated in future work through user studies or perceptual experiments to further strengthen the physical meaning of CAQA and enhance its alignment with human perception. It should be noted that, as an initial work in resource allocation for ISAC-driven AIGC networks, this paper considers a basic scenario with one single ISAC device and one AIGC server. However, in practice, a single device and an AIGC server may suffer from limited processing capacity. Future work should focus on more practical ISAC-based AIGC networks with multiple ISAC devices and AIGC servers, where collaborative sensing and coordinated generation across AIGC-enabled edge computing nodes would be the main challenge.

\bibliography{IEEEabrv,refbib}
\bibliographystyle{IEEEtran}

\begin{IEEEbiography}[{\includegraphics[width=1in,height=1.25in, clip,keepaspectratio]{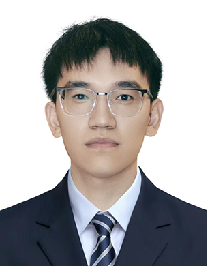}}]{Ningzhe Shi} (Graduate Student Member, IEEE) received the B.S. degree in communication engineering from Xidian University, Xi'an, China, in 2021. He is currently working toward the Ph.D. degree with the Institute of Computing Technology, Chinese Academy of Sciences, Beijing, China. His research interests include mobile edge computing, reinforcement learning, and next-G technologies such as integrated sensing, computation, and communication (ISCC). He has also contributed as a reviewer for several peer-reviewed journals and international conferences.\end{IEEEbiography}

\begin{IEEEbiography}[{\includegraphics[width=1in,height=1.25in, clip,keepaspectratio]{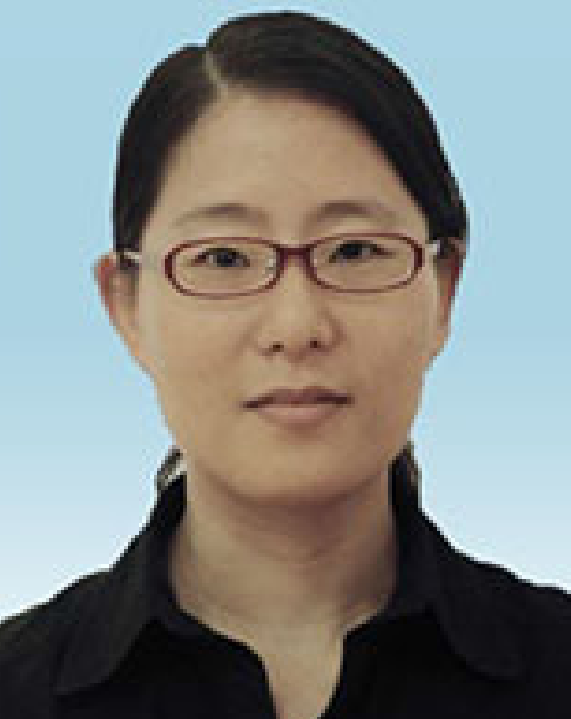}}]{Yiqing Zhou} (Senior Member, IEEE) received the B.S. degree in communication and information engineering and the M.S. degree in signal and information processing from Southeast University, China, in 1997 and 2000, respectively, and the Ph.D. degree in electrical and electronic engineering from the University of Hong Kong, Hong Kong, in 2004. She is currently a Professor with the Wireless Communication Research Center, Institute of Computing Technology, Chinese Academy of Sciences. She has published over 150 articles and four books/book chapters in the areas of wireless mobile communications. She received best paper awards from WCSP2019, IEEE ICC2018, ISCIT2016, PIMRC2015, ICCS2014, and WCNC2013. She also received the 2014 Top 15 Editor Award from IEEE TVT and the 2016-2017 Top Editors of ETT. She is also the TPC Co-Chair of ChinaCom2012, an Executive Co-Chair of IEEE ICC2019, a Symposia Co-Chair of ICC2015, a Symposium Co-Chair of GLOBECOM2016 and ICC2014, a Tutorial Co-Chair of ICCC2014 and WCNC2013, and the Workshop Co-Chair of SmartGridComm2012 and GlobeCom2011. She is also the Associate/Guest Editor of IEEE Internet of Things Journal, IEEE Transactions on Vehicular Technology (TVT), IEEE Journal on Selected Areas in Communications (JSAC) (Special issue on ``Broadband Wireless Communication for High Speed Vehicles" and ``Virtual MIMO"), Transactions on Emerging Telecommunications Technologies) (ETT), and Journal of Computer Science and Technology (JCST).\end{IEEEbiography}

\begin{IEEEbiography}[{\includegraphics[width=1in,height=1.25in, clip,keepaspectratio]{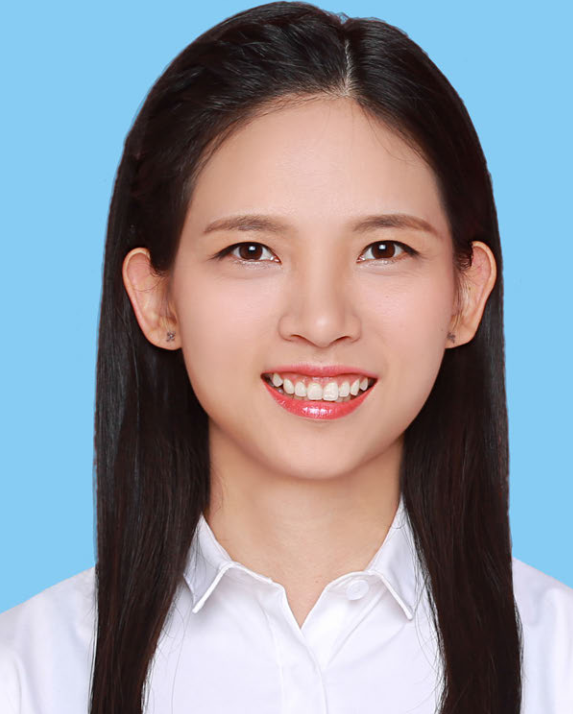}}]{Ling Liu} (Member, IEEE) received the B.S. degree in communication engineering from Nanchang University in 2012 and the Ph.D. degree in computer science and technology from the University of Chinese Academy of Sciences in 2018. She is currently an Associate Professor with the Wireless Communication Research Center, Institute of Computing Technology, Chinese Academy of Sciences. Her research interests include interference and resource management in ultra-dense networks and the convergence of communication, computing, and caching. She has also served as a reviewer for a number of refereed journals and international conferences. She has received the Best Paper Award from the IEEE ICC 2018.\end{IEEEbiography}

\begin{IEEEbiography}[{\includegraphics[width=1in,height=1.25in, clip,keepaspectratio]{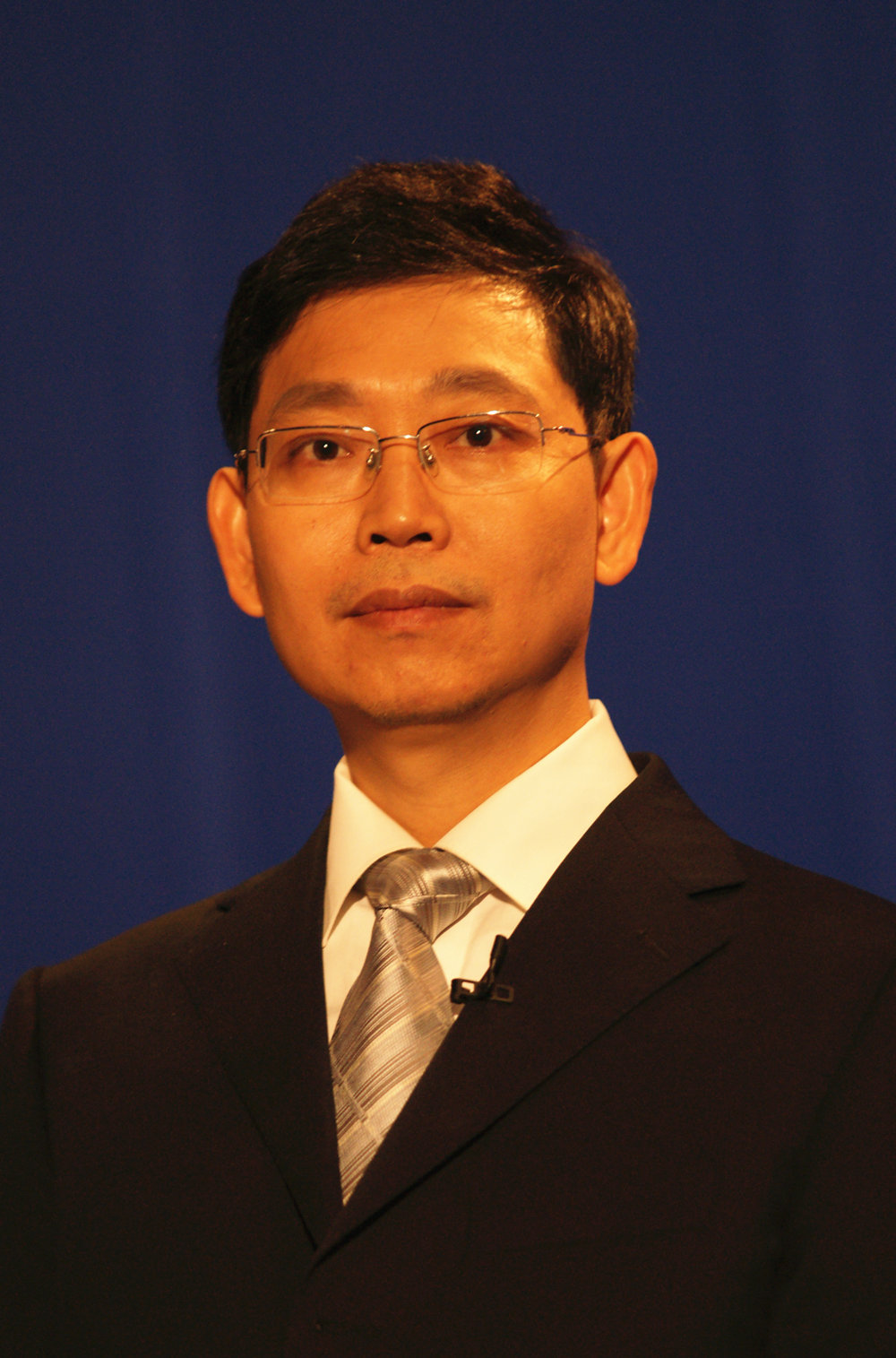}}]{Jinglin Shi} is currently the director of the Wireless Communication Technology Research Center, ICT/CAS. He has published 2 books and more than 100 papers in telecommunications journals and conference proceedings, and has more than 30 patents granted. His research interests include wireless communication system architecture, signal processing, and baseband processor design. He was the General Co-Chair of ChinaCom’12, and a member of the TPC of IEEE WCNC, ICC, AusWireless2006, ISCIT2007, and ChinaCom 2007 and 2009.\end{IEEEbiography}

\begin{IEEEbiography}[{\includegraphics[width=1in,height=1.25in,clip,keepaspectratio]{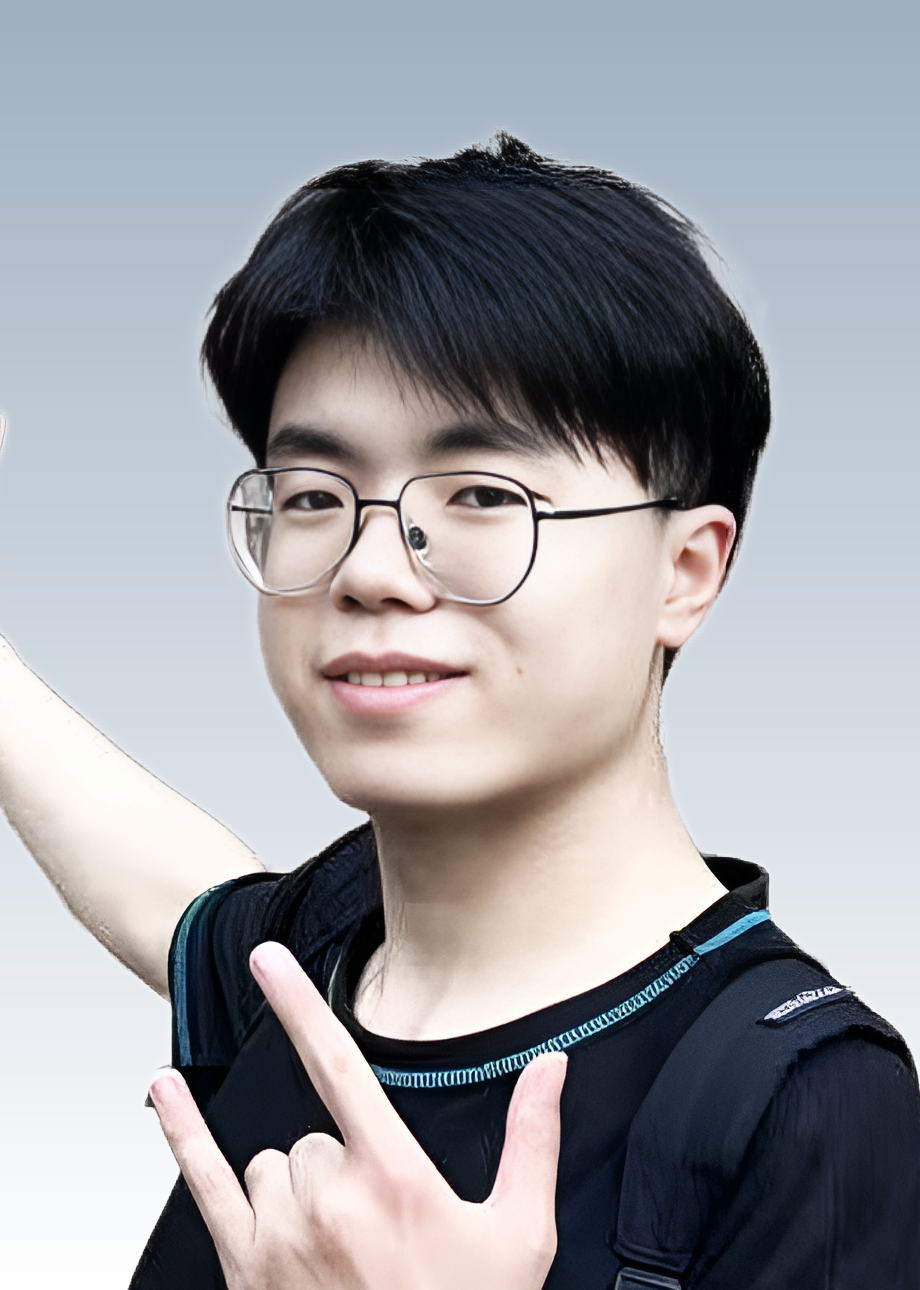}}]{Yihao Wu} received the B.S. degree in communication engineering from University of Science and Technology Beijing, in 2022. He is currently a Ph.D. candidate in Computer Science and Technology at the Wireless Communication Research Center, Institute of Computing Technology, Chinese Academy of Sciences. His research focuses on integrated sensing and communication, generative AI, and edge intelligence.\end{IEEEbiography}

\begin{IEEEbiography}[{\includegraphics[width=1in,height=1.25in, clip,keepaspectratio]{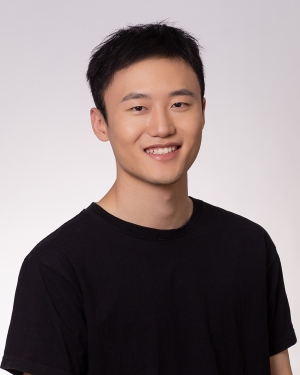}}]{Haiwei Shi} received the B.Eng. degree in communication engineering from Jilin University, Changchun, China, in 2022, and the M.S. degree in Computer Science and Technology from ShanghaiTech University, Shanghai, China, in 2025. He is currently pursuing the Ph.D. degree at the Institute of Computing Technology, Chinese Academy of Sciences, Beijing, China. His research interests include semantic communications and intelligent physical-layer signal processing.\end{IEEEbiography}

\begin{IEEEbiography}[{\includegraphics[width=1in,height=1.25in, clip,keepaspectratio]{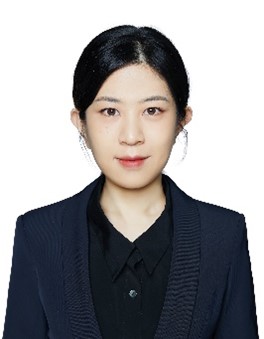}}]{Hanxiao Yu} (Member, IEEE) received the B.S. and Ph.D. degrees from Beijing Institute of Technology, Beijing, China, in 2015 and 2021, respectively. She is currently an Associate Research Fellow with the Wireless Communication Research Center, Institute of Computing Technology, Chinese Academy of Sciences, Beijing. She was a Postdoctoral Fellow with the School of Electronic and Information, Beijing Institute of Technology. Her research interests are in the area of 5G/6G communication system, LEO communication system, and multiple access technology.\end{IEEEbiography}
\vspace{11pt}

\vfill
\clearpage
\appendices
\renewcommand{\thesection}{\appendixname~\Alph{section}}
\section{Pseudocode for LPDRL-F Algorithm}
\begin{algorithm}[ht]
\caption{LPDRL-F for CAQA-AIGC Problem}
\label{alg:LPDRL}

Initialize actor, critic, and target critic parameters $\phi$, $\theta_j$, $\theta'_j$ for $j \in \{1,2\}$, and replay buffer $\Psi$.

\For{each episode to max episode}{
  \For{iteration $i=1$ to max iteration}{
    Collect current state $s_i$ and select action $a_i$ from the current policy.\par
    Decode $a_i$ to obtain the generating steps $\mathbf{z}(i)$ and the original sensing energy allocation.\par
    Apply the action filter in (10) to obtain the sensing energy allocation $\mathbf{E}_s(i)$.\par
    Run the RCE algorithm to obtain the communication energy allocation $\mathbf{E}_c(i)$.\par
    Compute reward $r_i$ using (11) and observe next state $s_{i+1}$.\par
    Store transition $\{s_i, a_i, r_i, s_{i+1}\}$ into $\Psi$.\par

    \If{$|\Psi| > \Psi_{\min}$}{
      Sample a batch $\Psi_b$ from $\Psi$.\par
      Update the critic network via (13).\par
      Update the actor network via (14).\par
      Update the target critic network via (15).\par
      Update the temperature parameter $\rho$ via $\nabla_{\rho} J(\rho)$~[31].\par
    }
  }
}
\end{algorithm}

\section{Pseudocode, Optimality Proof, and Complexity for RCE Algorithm}
\label{appendix:B}

\begin{algorithm}[ht!]
\caption{RCE Algorithm for ComRA Sub-problem}
\label{alg: RCE for ComRA}

Step 0: Given sensing energy and generating step solutions $\left\{ \mathbf{E}_s, \mathbf{z} \right\}$, and the communication energy bounds $E_{k,c}^{\min}$, $E_{k,c}^{\max}$, and remaining energy $E_r$.

\If{$E_{r} \leq 0$ or $E_{k,c}^{\max} < E_{k,c}^{\min}$ for $\forall k \in {\rm{{\cal K}}}$}
{
    Step 1: Set $E_{k,c} = 0$ for all users.
}
\Else
{
    Step 1: Sort users in descending order by priority metric
    \[
    \Lambda_k = \frac{\Theta_k(E_{k,s},z_k) B \log_2\left(1 + \frac{g_k P_c}{\delta^2} \right)}{K \beta P_c D_c}.
    \]

    \If{$E_r \leq \sum\nolimits_{k=1}^K E_{k,c}^{\min}$}
    {
        Step 2: Allocate $E_{k,c}^{\min}$ sequentially in the sorted order until the energy budget is exhausted.
    }
    \Else
    {
        Step 2: Allocate $E_{k,c}^{\min}$ to all users. \\
        Step 3: Allocate the additional remaining energy $E_r - \sum_{k=1}^K E_{k,c}^{\min}$ to users in sorted order, subject to $E_{k,c} \leq E_{k,c}^{\max}$.
    }
}
\end{algorithm}

\textbf{Optimality proof}: The communication energy allocation is formulated as a standard LP problem. The remaining energy ${E_r}$ is first allocated to meet each user’s minimum communication energy requirement. If additional energy ${E_r} - \mathop \sum \nolimits_{k = 1}^K E_{k,c}^{\min }$ remains, it is allocated in descending order of priority metric ${\Lambda _k}$, representing the CAQA achievable per unit of energy for user $k$. This ensures that users who contribute the most to the overall CAQA receive energy first. Each user is allocated up to their maximum communication energy limit $E_{k,c}^{\max }$. Once this limit is reached, the next user in the order is considered, and the process continues until all energy is allocated or all users have reached their maximum communication energy.

\textbf{Complexity Analysis}: The RCE algorithm consists of two main steps. (1) \textit{Priority Metric-Based Ranking}: Users are sorted based on the priority metric ${\Lambda_k}$, which reflects the communication efficiency per unit energy. This sorting step has a time complexity of $O(K \log K)$. (2) \textit{Communication Energy Allocation}: Steps 2 and 3 of Algorithm~\ref{alg: RCE for ComRA} perform a linear-time allocation process with complexity $O(K)$, where each user is first assigned their minimum required energy, and any remaining energy is distributed to higher-priority users, subject to their maximum limits. Therefore, the overall complexity of the RCE algorithm is $O(K \log K)$.

\end{document}